\documentclass[11pt]{article}

\usepackage[final]{acl}

\usepackage{times}
\usepackage{latexsym}
\usepackage[T1]{fontenc}
\usepackage{enumitem}
\usepackage[utf8]{inputenc}
\usepackage{graphicx} 
\graphicspath{{pic/}}
\usepackage{microtype}
\usepackage{inconsolata}

\usepackage{graphicx}
\usepackage{amsmath}
\definecolor{mintgreen}{RGB}{240, 255, 240}  
\definecolor{salmon}{RGB}{255, 240, 240}     

\definecolor{darkgreen}{RGB}{0, 100, 0}      
\definecolor{darkred}{RGB}{139, 0, 0}        

\usepackage{tabularx}
\usepackage{array}

\usepackage{booktabs}
\usepackage{adjustbox}
\usepackage[table]{xcolor}
\usepackage{pifont}
\usepackage{algorithm}
\usepackage{algpseudocode}
\newcommand{\cmark}{\textcolor{green!60!black}{\ding{51}}}
\newcommand{\xmark}{\textcolor{red!70!black}{\ding{55}}}
\newcommand{\pmark}{\textcolor{orange!80!black}{$\triangle$}}

\usepackage[most]{tcolorbox}
\usepackage{xcolor}
\usepackage{ragged2e}

\definecolor{promptblue}{HTML}{2F80ED}
\definecolor{promptgray}{HTML}{666666}

\usepackage{colortbl}  
\usepackage[table,xcdraw]{xcolor}  

\newcommand{\hmark}{\textcolor{blue}{\ding{51}\rotatebox[origin=c]{-6.2}{\kern-0.7em\ding{55}}}}

\title{RealWorldShop: Benchmarking and Improving Conversational Shopping Agents in Real-World E-commerce}

\author{
Xinwei Yang$^{1,2,4}$\thanks{This work was done during Xinwei Yang's internship at JD.com.} \quad 
Kelong Mao$^{1}$\quad 
Yudong Guo$^{1}$\quad  
Sulong Xu$^{1}$\quad 
Simiu Gu$^{1}$\\
\textbf{Chen Huang}$^{2,3,4}$\thanks{Corresponding author.} \quad 
\textbf{Wenqiang Lei}$^{2,4}$ \\
$^{1}$ JD.com\\
$^{2}$ College of Computer Science, Sichuan University  \\
$^{3}$ Institute of Data Science, National University of Singapore  \\
$^{4}$ Engineering Research Center of Machine Learning and Industry Intelligence, \\ Ministry of Education, China \\
\texttt{xinwei\_yang@stu.scu.edu.cn} \quad \texttt{huangc.scu@gmail.com} 
}

\begin{document}
\maketitle

\begin{abstract}
Large language models are reshaping e-commerce from static recommenders into interactive shopping assistants, yet real-world shopping requires session-level decision support: users reveal and revise constraints, coordinate multiple goals, and expect product-grounded recommendations over a full conversation. Existing benchmarks are mostly outcome-oriented or execution-oriented, leaving this evolving decision process under-evaluated. We introduce \textsc{RealWorldShop}, a benchmark built on 3.28M grounded products, structured shopping episodes, a profile-grounded and action-controlled user simulator, and role-play evaluation. Our analysis shows that current systems produce locally plausible responses but struggle with state tracking, constraint updating, and grounded convergence, especially under ambiguous intent, bundle, and multi-intent scenarios. We further propose \textsc{RealShop\_Agent}, an executable session-control framework with explicit state management, shopping-flow control, catalog-grounded retrieval, and runtime guards. Experiments show that \textsc{RealShop\_Agent} consistently outperforms strong baselines on \textsc{RealWorldShop}.
\end{abstract}

\section{Introduction}

Large language models are reshaping e-commerce systems from static recommenders into interactive shopping assistants that support purchase decisions across multi-turn conversations \cite{wang2026shoppingbench,lyu2025deepshop,liu2026recoworld}. However, real-world shopping is not simply a matter of returning a relevant item. Users often begin with underspecified intents, reveal constraints gradually, revise earlier preferences, compare alternatives, and sometimes shift purchase targets \cite{kim2024stop,huang2026towards,wang2025opera}. Valid recommendations must also be grounded in real catalog evidence, including category, price, attributes, compatibility, and fulfillment constraints \cite{li2026rectom,chen2025recusersim,li2026comboshoppingbench}. Consequently, real-world shopping agents must elicit hidden needs, maintain and update user state, coordinate subgoals, and guide the interaction toward a grounded and actionable purchase throughout the interaction session.

Despite growing interest in conversational shopping agents, existing benchmarks remain insufficient for evaluating agent performance in real-world shopping environments. Specifically, outcome-oriented benchmarks primarily evaluate preference elicitation or final-item relevance, but often rely on simplified product spaces, static intents, or limited state transitions \cite{huang2024concept,qin2024beyond,wang2025search}. In contrast, execution-oriented benchmarks emphasize filtering, sorting, and tool-use behaviors, while largely overlooking the conversational decision process behind those actions \cite{wang2026shoppingbench,cheng2026chatshopbuddy}. Consequently, current evaluation practice provides limited insight into critical real-world capabilities, such as hidden-constraint elicitation, state updating, catalog-grounded recommendation, bundle and multi-intent coordination, and deciding when to clarify, compare, recommend, or confirm \cite{du2025sapient,chhetri2025framework,zhao2026ecomstage,li2026rectom}.

In this work, we introduce \textsc{RealWorldShop}, a real-world benchmark for conversational shopping agents with session-level evaluation. \textsc{RealWorldShop} operationalizes realistic shopping interactions through four coordinated components. First, a \underline{3.28M}-product inventory grounds recommendations in real catalog evidence, including prices, categories, attributes, descriptions, and fulfillment signals. Second, approximately \underline{1,200} diverse shopping scenarios instantiate realistic decision contexts, covering explicit and ambiguous intents, single- and multi-subgoal sessions, bundle construction, upgrades, repairs, restocking and urgency-driven purchases. Third, more than \underline{2,000} synthesized user profiles support a profile-grounded and action-controlled user simulator that gradually exposes hidden intents and constraints, enabling controlled evaluation under partial observability. Finally, a session-level role-play protocol evaluates local response quality, process quality, and final convergence, distinguishing successful purchase decisions from early exits and max-turn failures. Together, these components enable the evaluation of conversational shopping agents under realistic shopping dynamics.

Using \textsc{RealWorldShop}, \textbf{we find that current shopping agents are substantially better at turn-level response generation than at session-level decision control.} Many systems produce fluent and plausible responses at individual turns, but their performance drops when they must preserve evolving user state, revise constraints after updates, coordinate multiple subgoals, and converge to a grounded purchase decision. These failures are especially pronounced in ambiguous-intent, bundle, multi-intent, and intent-shift scenarios, where hidden needs and changing constraints must be propagated across the full trajectory. These findings suggest that realistic conversational shopping should be evaluated under partial observability, catalog grounding, and convergence pressure, rather than treated as single-turn recommendation or isolated tool execution.

To address these failure modes, we propose \textsc{RealShop\_Agent}, an executable session-control framework for conversational shopping. Rather than treating shopping conversations as open-ended response generation, \textsc{RealShop\_Agent} models them as state-driven decision processes: it maintains evolving session state, decides when to clarify, retrieve, verify, compare, recommend, or confirm, grounds product decisions in catalog evidence, and applies runtime guards against over-retrieval, unsupported SKU-level claims, and stale-state recommendations. These capabilities are implemented through four executable modules: \textbf{Session State Manager}, \textbf{Shopping-Flow Controller}, \textbf{Catalog-Grounded Retriever}, and \textbf{Runtime Execution Guards}. Experiments on \textsc{RealWorldShop} show that \textsc{RealShop\_Agent} consistently outperforms strong baselines, and improves the success rate from 0.71 to 0.82 over the strongest baseline. These results suggest that robust shopping agents require explicit session-level control and grounded decision making beyond local response generation. Our contributions are threefold:

\begin{itemize}[leftmargin=*, itemindent=0.05cm, itemsep=-2pt]
    \item We introduce \textsc{RealWorldShop}, a session-level benchmark grounded in a 3.28M-SKU inventory, structured episodes, an action-controlled user simulator, and role-play evaluation.

    \item We propose \textsc{RealShop\_Agent}, an executable session-control framework with state management, shopping-flow control, catalog-grounded retrieval, and runtime guards.

    \item We show that current systems struggle with evolving session-level constraints, while \textsc{RealShop\_Agent} achieves the strongest overall performance on \textsc{RealWorldShop}.
\end{itemize}

\begin{figure*}[t]
    \centering
    \includegraphics[width=0.98\textwidth]{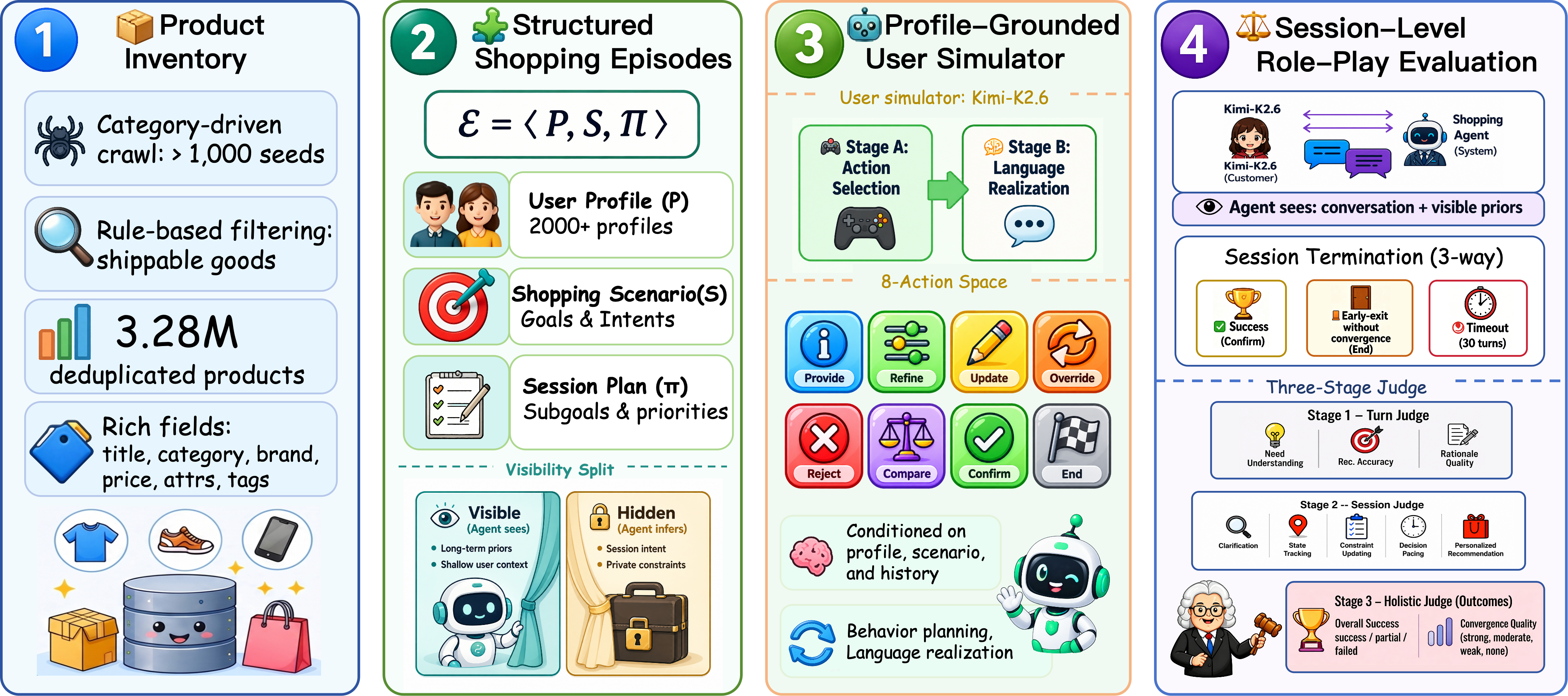}
    \caption{Overview of \textsc{RealWorldShop}. The benchmark integrates a 3.28M-SKU product inventory, approximately 1,200 structured shopping scenarios, and over 2,000 synthesized user profiles to construct profile-grounded user simulators, together with a session-level role-play protocol for evaluating conversational shopping agents.}
    \label{fig:overall_benchmark}
\end{figure*}

\section{Related Work}

\noindent\textbf{Conversational Shopping Agent Benchmark.}
Existing benchmarks for conversational shopping agents mainly follow two lines. 
\emph{Outcome-oriented} benchmarks evaluate preference elicitation and final-item relevance \cite{liang2024llm,chhetri2025framework,huang2024concept,kim2024stop,vlachou2025fashion,chen2025recusersim}, while \emph{execution-oriented} benchmarks test shopping actions, tool calls, or task-stage completion \cite{liu2026recoworld,lyu2025deepshop,zhao2026ecomstage,wang2025opera}. 
However, real-world shopping requires agents to operate over full sessions, where users reveal needs gradually, revise constraints, reject candidates, compare alternatives, and shift or combine purchase goals. 
\textsc{RealWorldShop} addresses this gap by evaluating complete multi-turn shopping trajectories under catalog grounding and partial observability, with metrics covering turn-level response quality, session-level process quality, holistic outcomes, and early-exit behavior. 
A detailed comparison with representative benchmarks is provided in Table~\ref{tab:benchmark_comparison}.

\noindent\textbf{Improving Conversational Shopping Agents.}
Recent methods improve conversational shopping agents through user profiling, strategic action selection, feedback modeling, tool-augmented reasoning, reinforcement learning, and recommendation representations \cite{shi2025personax,kim2025towards,cai2025agentic,zhang2026recthinker,cheng2026chatshopbuddy,zhao2025reason,zhang2026unleashing}. 
These advances are complementary to our setting and provide useful building blocks for more capable shopping agents. 
However, they less directly operationalize conversational shopping as a session-control problem, where an agent must maintain evolving user state, decide when to clarify, retrieve, verify, or recommend, ground product decisions in current catalog evidence, and prevent responses that conflict with revised constraints or unsupported product claims. 
\textsc{RealShop\_Agent} addresses this perspective through four executable modules: a Session State Manager, a Shopping-Flow Controller, a Catalog-Grounded Retriever, and Runtime Execution Guards.

\begin{table}[t]
\centering
\small
\setlength{\tabcolsep}{4pt}
\renewcommand{\arraystretch}{1.10}
\begin{tabularx}{\linewidth}{@{}lX@{}}
\toprule
\textbf{Component} & \textbf{Illustrative Episode} \\
\midrule

$P$ 
& A budget-conscious, low-patience user preferring compact products, with latent constraints on counter space, noise, and household size. \\

$S$ 
& Ambiguous bundle shopping for a small-apartment kitchen, starting from a vague request for ``something useful for my kitchen.'' \\

$\Pi$ 
& The simulator reveals space constraints, rejects an oversized appliance, updates the budget, and requests a compatible bundle. \\

\textit{Target} 
& Clarification, state tracking, constraint updating, bundle coordination, and grounded convergence. \\

\bottomrule
\end{tabularx}
\caption{Shopping episode illustration.}
\label{tab:running_example}
\end{table}

\section{RealWorldShop Benchmark}

Figure~\ref{fig:overall_benchmark} illustrates the construction and evaluation pipeline of \textsc{RealWorldShop}. 
Rather than treating conversational shopping as single-turn recommendation, \textsc{RealWorldShop} evaluates agents as session-level decision makers through four components: 
(1) a real-product inventory for catalog-faithful grounding, 
(2) structured shopping episodes for realistic and evolving shopping contexts, 
(3) a profile-grounded user simulator for controlled multi-turn interaction, and 
(4) a role-play evaluation protocol for measuring turn quality, session process, and final convergence.

\subsection{Product Inventory}
\label{sec:benchmark-inventory}
We construct the product pool from authorized category-driven crawls over a large real-world e-commerce platform. 
After rule-based filtering and deduplication, the final inventory contains 3.28M deduplicated products across major product groups. 
Each record retains catalog fields such as title, price, attributes, category labels, descriptions, and fulfillment signals, enabling catalog-faithful grounding. 
These fields allow the benchmark to verify whether recommendations are not only preference-matched, but also faithful to real product evidence with respect to category, price, attributes, and fulfillment constraints. 
Further details are provided in Appendix~\ref{app:benchmark-inventory}.

\subsection{Shopping Episode Construction}
\label{sec:benchmark-episodes}

Each evaluation instance in \textsc{RealWorldShop} is a structured shopping episode 
$E=\langle P,S,\Pi\rangle$, where $P$ specifies the user profile, $S$ defines the shopping scenario, 
and $\Pi$ controls how the session unfolds. 
This formulation separates the latent shopping state from the agent-visible context, allowing the benchmark to test whether an agent can elicit, maintain, and update user needs through interaction.
Episodes are synthesized through a category-anchored pipeline rather than as fixed dialogues. 
Starting from product category queries and source paths, the pipeline instantiates task archetypes such as bundle construction, upgrades, repairs, restocking, mixed errands, and urgency-driven purchases, together with constraints involving compatibility, budget, risk, logistics, and bundle splitting. 
The session plan $\Pi$ specifies whether the intent is explicit or ambiguous, whether one or multiple subgoals are involved, and when hidden constraints, preference changes, rejections, or intent shifts are revealed. 
This enables controlled evaluation of clarification, state tracking, constraint updating, and multi-subgoal coordination. 
Table~\ref{tab:running_example} provides an example, with further details in Appendix~\ref{app:benchmark-episodes}.

\subsection{User Profile Construction}
\label{sec:benchmark-profile}

Each shopping episode is grounded in a structured user profile $P$ covering long-term preferences, historical behavior, current intent, constraints, interaction style, and optional multi-intent fields. 
We split each profile into an agent-visible prior $\tilde{P}$ and a latent state $P \setminus \tilde{P}$, so that agents must infer task-specific constraints through dialogue rather than assuming fully observed preferences.
Profiles cover both explicit-intent cases, where the goal is directly searchable from the initial request, and ambiguous-intent cases, where goals and constraints are revealed gradually. 
The schema distinguishes hard, soft, and negotiable requirements over budget, compatibility, risk tolerance, logistics, brand preference, and interaction style. 
To improve grounding, profile synthesis can be conditioned on local catalog evidence, and unsupported constraints are downgraded to soft or negotiable preferences. 
All profiles are normalized and validated for schema consistency, multi-intent structure, ambiguity fields, category anchoring, and catalog grounding. 
Further details are provided in Appendix~\ref{app:benchmark-profile}.

\subsection{Evaluation Protocol}
\label{sec:benchmark-protocol}
We evaluate conversational shopping systems through a role-play-based, session-level protocol. 
The protocol separates three roles: the simulator generates user behavior from the hidden episode state, the evaluated agent responds using only visible context and dialogue history, and the judge evaluates the completed trajectory.

\paragraph{Role-play setup.}
For each episode, Kimi-K2.6 plays the customer, conditioned on the structured episode configuration and dialogue history. 
The shopping agent, i.e., the system under evaluation, observes only the conversation history and the visible prior $\tilde{P}$, and must progressively infer hidden intents and constraints through interaction.

\paragraph{Interaction and termination.}
Conversations unfold turn by turn: the simulator first generates a user utterance under its action-controlled policy, and the agent then responds. 
A session terminates in one of three states: \textsc{Success}, where the interaction reaches a grounded and actionable purchase decision; \textsc{Early-Exit-Without-Convergence}, where the user ends the session before committing; or \textsc{Max-Turn Failure}, where the maximum of 30 assistant turns is reached. 
This taxonomy separates voluntary abandonment from capacity-limited non-convergence.

\paragraph{Three-level Judging \& Evaluation Reliability.}
After each episode, a three-stage GPT-5.5 judging pipeline evaluates the dialogue, tool trace, and episode configuration. The turn-level judge measures local response quality, including need understanding, recommendation accuracy, and rationale quality. The session-level judge measures process quality, including clarification, state tracking, constraint updating, decision pacing, and personalization. Finally, the holistic judge determines whether the session reaches an actionable purchase decision and assigns \texttt{overall\_success} and \texttt{convergence\_quality}. All turn- and session-level dimensions are scored on a $\{0,1,2\}$ scale. We assess judge reliability using repeated judging passes and a stratified 150-session human validation subset, where automatic judgments achieve QWK/$\kappa$ scores of 0.76--0.89 and exact agreement rates of 0.88--0.97 across dimensions. Further details are provided in Appendix~\ref{app:judge-reliability}.


\begin{table*}[t]
\centering
\scriptsize
\setlength{\tabcolsep}{1.9pt}
\renewcommand{\arraystretch}{1.15}

\resizebox{\textwidth}{!}{%
\begin{tabular}{l *{14}{c}}
\toprule

& \multicolumn{4}{c}{\textbf{Turn-Level Quality} $\uparrow$} & \multicolumn{6}{c}{\textbf{Session-Level Process Quality} $\uparrow$} & \multicolumn{1}{c}{\textbf{Turn--Session}} & \multicolumn{2}{c}{\textbf{Holistic Outcome}} & \multicolumn{1}{c}{\textbf{User Signal}} \\

\cmidrule(lr){2-5}
\cmidrule(lr){6-11}
\cmidrule(lr){12-12}
\cmidrule(lr){13-14}
\cmidrule(lr){15-15}

\textbf{Model} & \textbf{Need} & \textbf{Rec. Acc.} & \textbf{Rationale} & \textbf{Turn Avg.} & \textbf{Clarif.} & \textbf{State} & \textbf{Constraint} & \textbf{Pace} & \textbf{Personal.} & \textbf{Sess. Avg.} & \textbf{Gap} $\downarrow$ & \textbf{Succ.} $\uparrow$ & \textbf{Conv.} $\uparrow$ & \textbf{Early Exit} $\downarrow$ \\

\midrule

GPT-5 \cite{singh2025openai} & \textbf{1.62} & \textbf{1.41} & \textbf{1.33} & \textbf{1.45} & \textbf{1.68} & \textbf{1.23} & \textbf{1.09} & \textbf{0.94} & \textbf{1.08} & \textbf{1.20} & 0.25 & \textbf{0.71} & \textbf{2.49} & \textbf{0.10} \\

Gemini-2.5-Flash \cite{comanici2025gemini} & 0.91 & 0.80 & 0.84 & 0.85 & 1.21 & 0.69 & 0.56 & 0.59 & 0.45 & 0.70 & \textbf{0.15} & 0.38 & 1.64 & 0.35 \\

GPT-4o \cite{hurst2024gpt} & 1.30 & 0.71 & 0.75 & 0.92 & 0.95 & 0.62 & 0.49 & 0.37 & 0.35 & 0.56 & 0.36 & 0.38 & 1.48 & 0.39 \\

DeepSeek-V3.2 \cite{liu2025deepseek} & 1.36 & 0.92 & 0.81 & 1.03 & 1.41 & 0.80 & 0.63 & 0.64 & 0.52 & 0.80 & 0.23 & 0.56 & 1.93 & 0.27 \\

Kimi-K2.6 \cite{team2025kimi} & \underline{1.45} & \underline{1.22} & \underline{1.21} & \underline{1.29} & \underline{1.54} & 0.90 & 0.88 & 0.70 & \underline{0.95} & \underline{0.99} & 0.30 & \underline{0.67} & \underline{2.23} & 0.16 \\

Qwen3.5-27B \cite{yang2025qwen3} & 1.41 & 0.83 & 0.80 & 1.01 & 1.49 & 0.58 & 0.57 & 0.50 & 0.36 & 0.70 & 0.31 & 0.55 & 1.74 & 0.25 \\

Qwen3.5-35B-A3B \cite{yang2025qwen3} & 1.42 & 0.78 & 0.79 & 1.00 & 1.30 & 0.62 & 0.46 & 0.37 & 0.43 & 0.64 & 0.36 & 0.53 & 1.66 & 0.31 \\

\midrule

\textit{LLM Avg.} & 1.35 & 0.95 & 0.93 & 1.08 & 1.37 & 0.78 & 0.67 & 0.59 & 0.59 & 0.80 & 0.28 & 0.54 & 1.88 & 0.26 \\

\midrule

\multicolumn{15}{l}{\textit{Specialized Methods (DS-V3.2 backbone)}} \\

AgentCF \cite{zhang2024agentcf} & 1.37 & 1.15 & 1.03 & 1.18 & 1.42 & \underline{0.97} & 0.87 & 0.61 & 0.76 & 0.93 & 0.26 & 0.61 & 2.15 & 0.23 \\

CRAVE \cite{zhu2025llm} & 1.38 & 1.01 & 0.89 & 1.09 & 1.43 & 0.79 & 0.83 & 0.57 & 0.80 & 0.88 & 0.21 & 0.57 & 1.98 & 0.23 \\

PersonaX \cite{shi2025personax} & 1.36 & 1.13 & 0.92 & 1.14 & 1.45 & 0.92 & \underline{0.89} & \underline{0.71} & 0.86 & 0.97 & \underline{0.17} & 0.60 & 2.14 & \underline{0.15} \\

CSI \cite{kim2025towards} & 1.25 & 1.18 & 0.93 & 1.12 & 1.43 & 0.71 & 0.70 & 0.63 & 0.82 & 0.86 & 0.26 & 0.59 & 1.94 & 0.24 \\

AFL \cite{cai2025agentic} & 1.29 & 1.19 & 0.96 & 1.15 & 1.33 & 0.77 & 0.81 & 0.69 & 0.70 & 0.86 & 0.29 & 0.62 & 1.90 & 0.21 \\

\midrule

\textit{Method Avg.} & 1.33 & 1.13 & 0.95 & 1.14 & 1.41 & 0.83 & 0.82 & 0.64 & 0.79 & 0.90 & 0.24 & 0.60 & 2.02 & 0.21 \\

\bottomrule
\end{tabular}%
}

\caption{
Overall benchmark results on \textsc{RealWorldShop}.
Turn Avg. is the arithmetic mean of Need, Rec. Acc., and Rationale,
while Sess. Avg. is the arithmetic mean of Clarif., State, Constraint,
Pace, and Personal.
Gap measures the degradation from turn-level quality to session-level
process quality and is computed as the unrounded Turn Avg. minus the
unrounded Sess. Avg.
Succ. denotes success rate, Conv. denotes convergence quality, and
Early Exit denotes the simulator-side early-exit rate.
All aggregate metrics are computed before rounding.
Best and second-best results across all evaluated systems are marked
in \textbf{bold} and \underline{underline}, respectively.
Tied results receive the same formatting.
}

\label{tab:main_evaluation_result}
\end{table*}

\section{Benchmark Analysis}
\subsection{Benchmark Setup}

\noindent\textbf{Overview}. All experiments are performed on RealWorldShop, utilizing the rigorous evaluation protocol and metrics defined in Section \ref{sec:benchmark-protocol}. 

\noindent\textbf{Baselines}. 
We evaluate two groups of baselines: general LLMs and recent shopping-agent methods. 
The LLM group includes \textit{GPT-5}~\cite{singh2025openai}, \textit{Gemini-2.5-Flash}~\cite{comanici2025gemini}, \textit{GPT-4o}~\cite{hurst2024gpt}, \textit{DeepSeek-V3.2}~\cite{liu2025deepseek}, \textit{Kimi-K2.6}~\cite{team2025kimi}, \textit{Qwen3.5-27B}~\cite{yang2025qwen3}, and \textit{Qwen3.5-35B-A3B}~\cite{yang2025qwen3}; all models are evaluated under the same task instructions, agent-visible inputs, tool interfaces, and judging pipeline, while their own generated responses and tool traces are evaluated independently.
The shopping-agent group includes \textit{AgentCF}~\cite{zhang2024agentcf}, \textit{CRAVE}~\cite{zhu2025llm}, \textit{PersonaX}~\cite{shi2025personax}, \textit{CSI}~\cite{kim2025towards}, and \textit{AFL}~\cite{cai2025agentic}.

\noindent\textbf{Evaluation Dataset}. 
We construct a held-out diagnostic set from \textsc{RealWorldShop} and evaluate all systems on the same episodes. 
Each episode is annotated along three axes: scenario type, interaction phenomenon, and interaction profile. 
For each category, we sample 100 episodes, covering six scenario types: explicit single-item, explicit bundle, ambiguous single-item, ambiguous bundle, multi-intent shopping, and long-horizon sessions; six interaction phenomena: constraint refinement, product comparison, constraint update, preference override, recommendation rejection, and bundle re-planning; and six user profiles: collaborative, high-patience, low-skepticism, efficiency-first, high-skepticism, and low-patience. 
Early-exit and non-exit sessions are grouped by realized outcomes rather than pre-defined sampling labels.
For each system, we repeat the full evaluation five times and report the mean across the five runs.

\begin{figure}[t]
    \centering
    \includegraphics[width=0.95\linewidth]{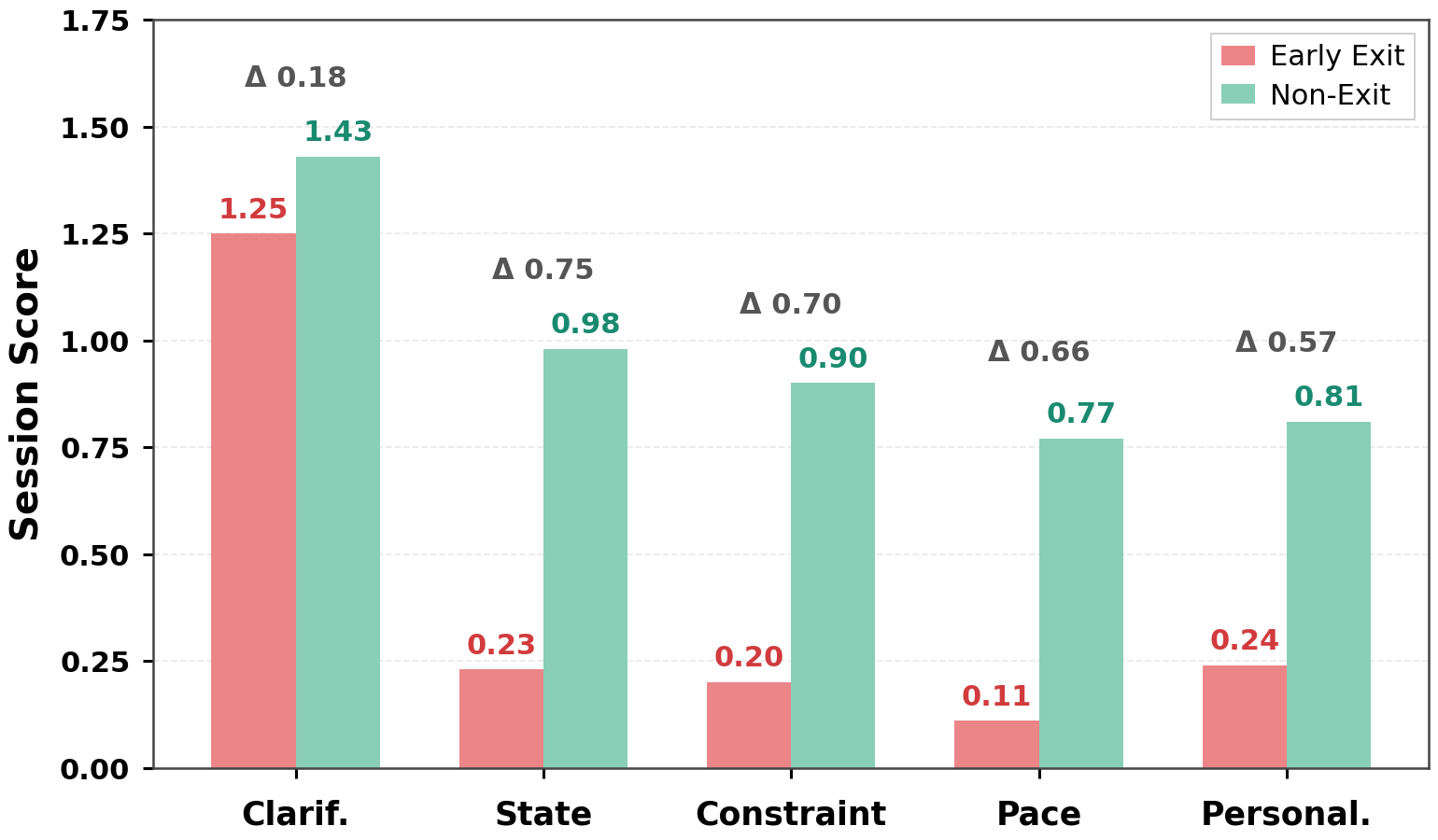}
    \caption{Early-exit sessions reveal weak session-level process control. Compared with non-exit sessions, early-exit sessions show larger degradation in state tracking, constraint updating, and decision pacing, indicating that user abandonment is closely tied to failures in maintaining and revising session state.}
    \label{fig:early_exit}
\end{figure}

\begin{figure}[t]
    \centering
    \includegraphics[width=0.99\linewidth]{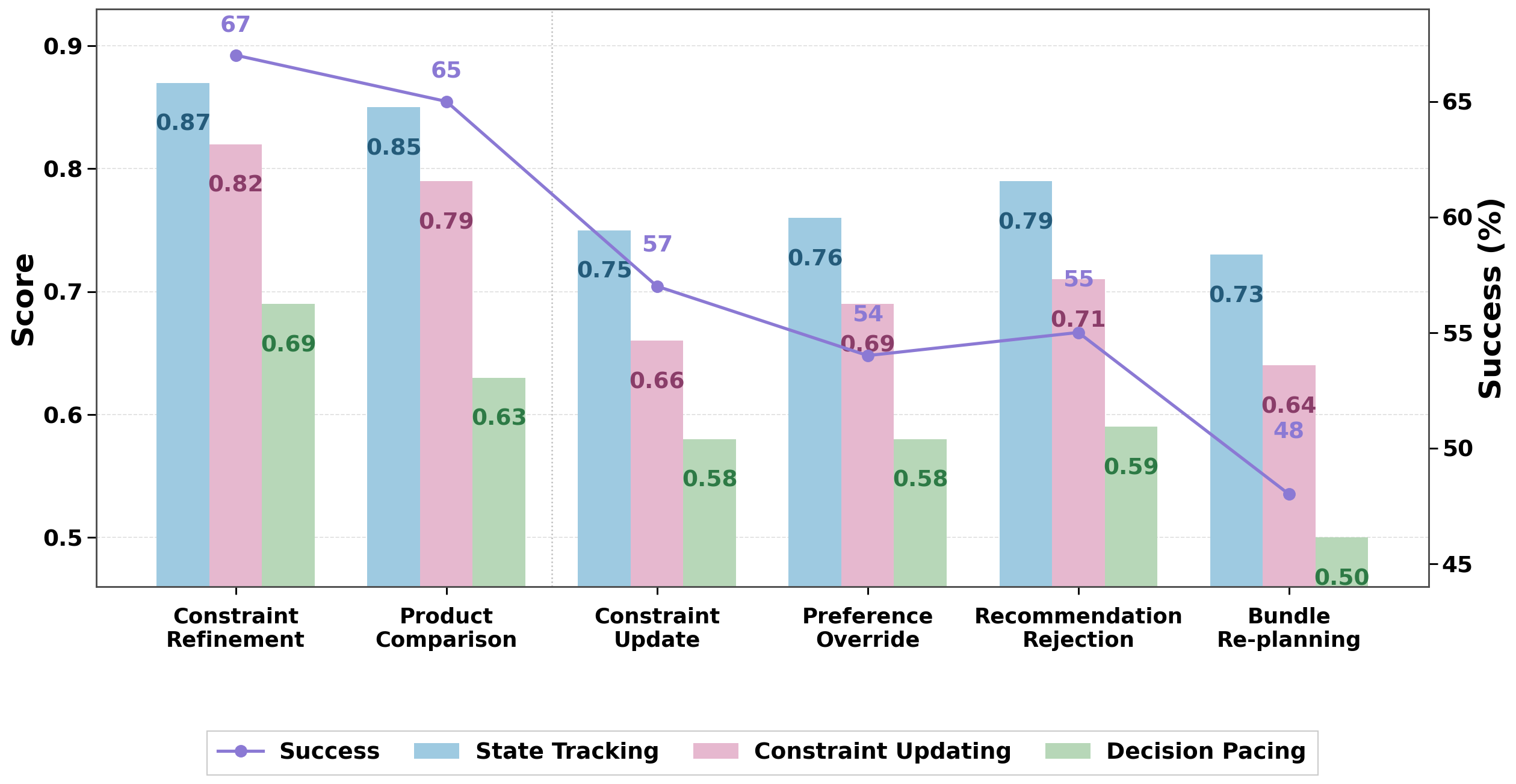}
    \caption{Interaction phenomena stress state revision and trajectory management. Performance declines under constraint updates, preference overrides, recommendation rejections, and bundle re-planning, showing that agents struggle to propagate revised user information through the full shopping trajectory.}
    \label{fig:interaction_phenomena}
\end{figure}

\begin{figure}[t]
    \centering
    \includegraphics[width=0.98\linewidth]{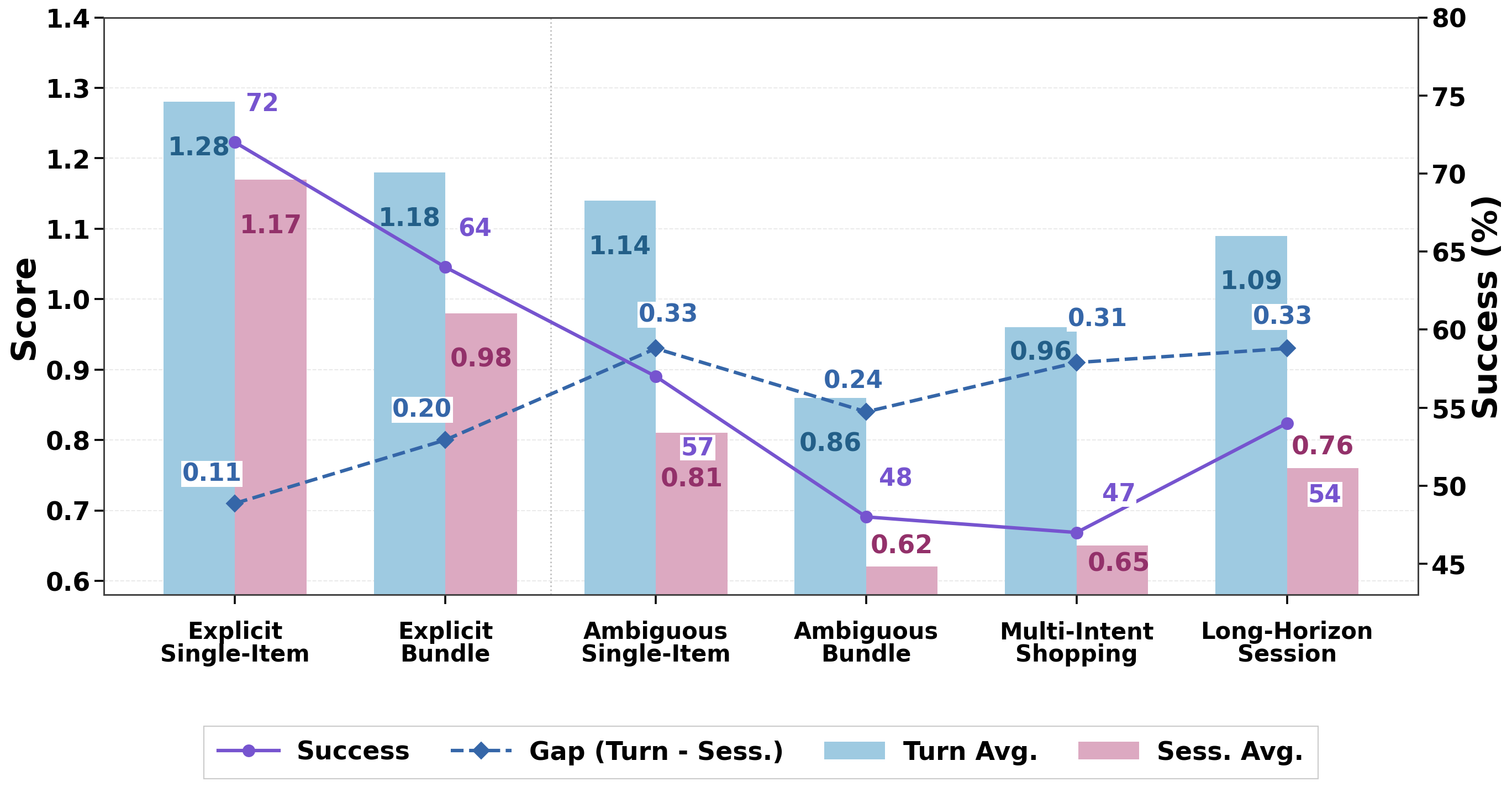}
    \caption{Scenario complexity amplifies the turn-to-session quality gap. Agents perform best in explicit single-item sessions, while ambiguous, bundle, multi-intent, and long-horizon scenarios expose larger gaps between local response quality and session-level shopping success.}
    \label{fig:across_scenarios}
\end{figure}

\begin{figure}[t]
    \centering
    \includegraphics[width=0.99\linewidth]{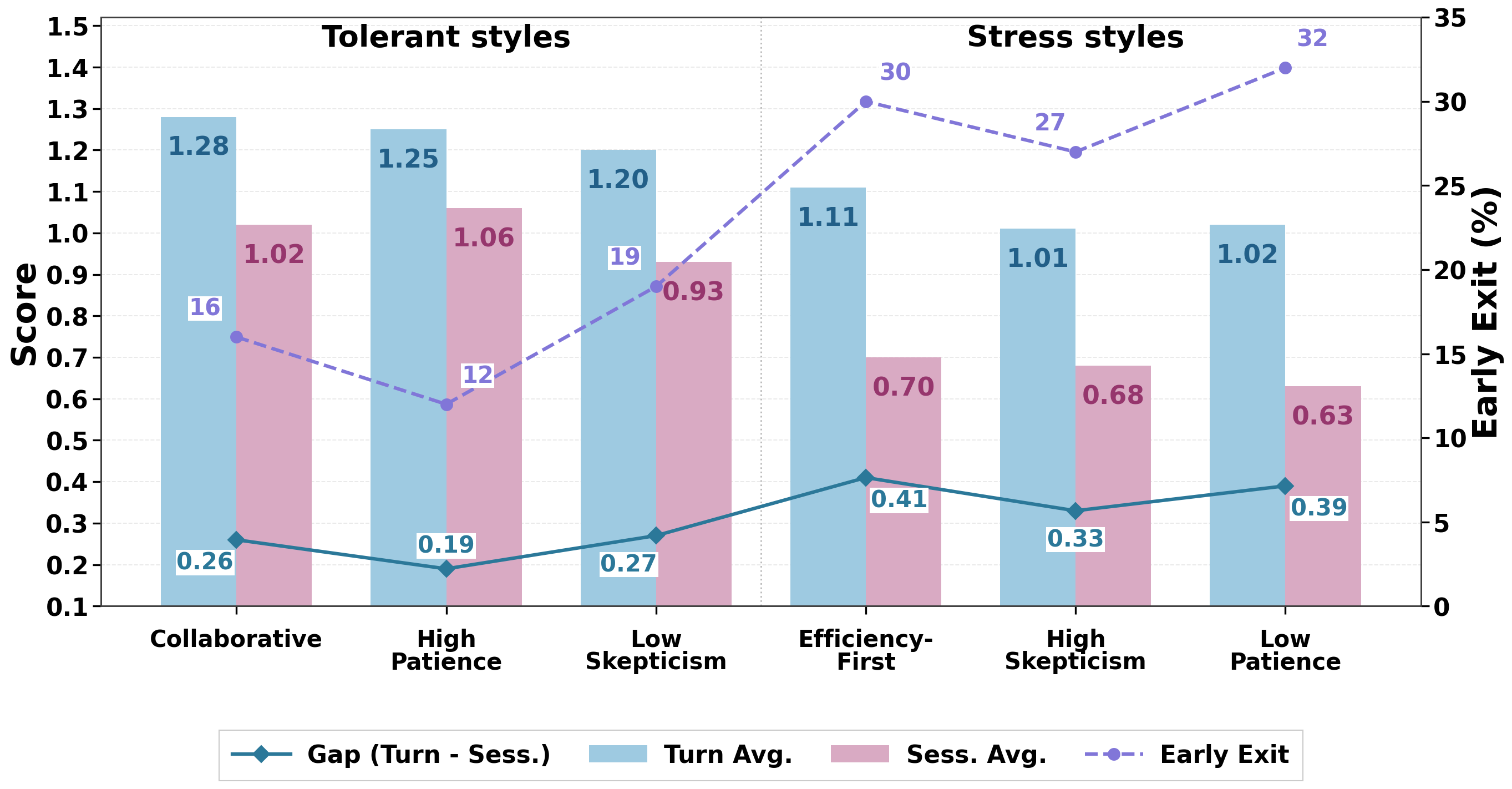}
    \caption{Demanding interaction styles increase session fragility. Efficiency-first, high-skepticism, and low-patience users lead to larger turn-to-session gaps and higher early-exit rates, highlighting the need to evaluate shopping agents under realistic convergence pressure.}
    \label{fig:interaction_style}
\end{figure}

\subsection{Benchmark Findings}
\label{sec:benchmark_finding}

\paragraph{Finding 1: High single-turn scores do not guarantee successful shopping sessions.}
Table~\ref{tab:main_evaluation_result} evaluates agents at two levels: 
\textit{Turn Avg.} captures local response quality, whereas \textit{Sess. Avg.} captures trajectory-level shopping competence, including state tracking, constraint updating, subgoal coordination, and grounded convergence. 
Across all systems, the session-level score is consistently lower than the turn-level score. 
The average general LLM drops from 1.08 to 0.80, and specialized shopping-agent methods drop from 1.14 to 0.90. 
Even GPT-5 shows the same gap, decreasing from 1.45 to 1.20. 
These results indicate that locally plausible responses do not guarantee coherent shopping behavior over an entire session. 
An agent may handle the current user utterance well, but still fail to maintain evolving user state, coordinate multiple needs, revise constraints, and converge to a grounded purchase decision. 
Thus, turn-level evaluation alone risks substantially overestimating real shopping-agent competence.

\paragraph{Finding 2: Early exits are driven more by state-tracking failures than by poor clarification.}
Figure~\ref{fig:early_exit} shows that early-exit sessions underperform non-exit sessions across session-level dimensions, but the degradation is highly uneven. 
Clarification drops only moderately from 1.43 to 1.25, whereas state tracking drops sharply from 0.98 to 0.23. 
This contrast suggests that user abandonment is not mainly caused by insufficient questioning; rather, users are more likely to exit when the agent fails to remember, update, or act on information that has already been revealed. Figure~\ref{fig:interaction_phenomena} further shows that failures become more severe when the dialogue requires explicit state revision. 
In particular, bundle re-planning is one of the most challenging cases, with success falling to 48\% and process scores dropping to 0.64 for constraint updating and 0.50 for decision pacing. 
These results indicate that current agents struggle to propagate revised information through the session state, candidate set, and downstream recommendation plan.

\paragraph{Finding 3: Ambiguity, long horizons, and demanding user styles expose failures hidden by turn-level metrics.}
Figure~\ref{fig:across_scenarios} shows that agents perform best in explicit single-item sessions, achieving a success rate of 72\% and the smallest local-to-session gap of 0.11. 
As scenarios become more complex, performance declines: ambiguous single-item sessions reduce success to 57\%, while ambiguous single-item and long-horizon sessions jointly exhibit the largest gap of 0.33.
This suggests that ambiguity, bundle construction, multi-intent coordination, and long-range dependencies amplify trajectory-level weaknesses that are not captured by turn-level quality alone. Figure~\ref{fig:interaction_style} further shows that demanding interaction styles intensify these failures. 
Low-patience, high-skepticism, and efficiency-first users reduce session-level scores to 0.63--0.70 and raise early-exit rates to 27\%--32\%, compared with 12\%--19\% under more tolerant styles. 
Therefore, realistic shopping evaluation should stress-test agents under partial observability, evolving constraints, catalog grounding, and convergence pressure. Additional diagnostic analyses supporting these findings are provided in Appendix~\ref{mored}, where we further examine early-exit behavior, interaction phenomena, scenario complexity, and user interaction styles.

\section{\textsc{RealShop\_Agent}}
\label{sec:realshop_agent}

\paragraph{Overview \& Problem Setup.}
The findings from \textsc{RealWorldShop} suggest that conversational shopping is better viewed as a session-level decision-control problem than as isolated recommendation. Users may gradually reveal or revise needs, reject suggestions, compare alternatives, and shift purchase targets. An effective shopping agent must therefore maintain evolving state, choose when to clarify, retrieve, verify, compare, recommend, or confirm, and ground product claims in up-to-date catalog evidence.

\textsc{RealShop\_Agent} is a diagnosis-guided session-control framework that addresses these challenges through four executable modules: a Session State Manager for evolving user state, a Shopping-Flow Controller for action timing, a Catalog-Grounded Retriever for evidence grounding, and Runtime Execution Guards for preventing unsupported or stale recommendations. Together, these modules form an explicit control loop for session-level shopping.

Following the episode formulation in Section~3, each shopping episode is represented as $E=\langle P,S,\Pi\rangle$, where $P$ is the structured user profile, $S$ the shopping scenario, and $\Pi$ the session plan. At turn $t$, the assistant observes only the visible prior $\widetilde{P}$ and dialogue history $h_{<t}$, while latent profile fields and future session dynamics remain hidden. \textsc{RealShop\_Agent} maintains an explicit session state $M_t$ and selects among six high-level actions: \textsc{Clarify}, \textsc{Retrieve}, \textsc{Verify}, \textsc{Compare}, \textsc{Recommend}, and \textsc{Confirm}. Depending on the selected action, it may invoke catalog tools $o_t\in\mathcal{T}$, update grounded evidence, and generate a response subject to grounding and state-consistency checks. Algorithmic details are provided in Appendix~\ref{app:realshop_agent_details}.

\subsection{M1: Session State Manager}

The Session State Manager maintains the evolving state $M_t$, covering revealed intents, active subgoals, constraints, rejected options, confirmed preferences, and stale dependencies. 
Since shopping sessions are non-stationary, newly revealed constraints override long-term preferences, and dependent subgoals are marked stale when budgets, candidates, compatibility requirements, or bundle anchors change. 
The updated $M_t$ guides downstream action selection and product grounding.

\subsection{M2: Shopping-Flow Controller}

The Shopping-Flow Controller maps the current session state $M_t$ to one of six high-level shopping actions: \textsc{Clarify}, \textsc{Retrieve}, \textsc{Verify}, \textsc{Compare}, \textsc{Recommend}, and \textsc{Confirm}. This module determines not only what information or evidence is needed, but also when the agent should progress from elicitation to search, verification, comparison, recommendation, and final commitment.
It selects \textsc{Clarify} when blocking information is missing; \textsc{Retrieve} when the active subgoal is sufficiently specified but fresh catalog evidence is unavailable; \textsc{Verify} when retrieved candidates have not yet been checked against active constraints; \textsc{Compare} when multiple viable candidates remain or the user requests trade-off analysis; \textsc{Recommend} when grounded candidates satisfy the active constraints and are aligned with the current subgoal; and \textsc{Confirm} when the user is ready to commit and the proposed recommendation or purchase plan is actionable.
This explicit action policy reduces premature recommendations, redundant questioning, unsupported comparisons, and delayed convergence.

\subsection{M3: Catalog-Grounded Retriever}

The Catalog-Grounded Retriever connects the state-driven control process to real product evidence. When retrieval or verification is triggered, it returns structured evidence records $E_t=\{e_i\}$, including SKU identifiers, category labels, attributes, prices, fulfillment signals, relevance scores, and retrieval metadata. This module is necessary because even a well-timed recommendation can be invalid if its product claims are not supported by the current catalog evidence. Therefore, concrete SKU-level claims must be traceable to the latest retrieved records, and evidence tied to stale intents, rejected options, or superseded constraints is invalidated before recommendation.

\subsection{M4: Runtime Execution Guards}
These Guards form the final consistency layer before response generation. 
Since agents may still over-retrieve, overstate product properties, or recommend candidates that conflict with updated constraints, \textsc{ToolBudget} limits redundant retrieval, \textsc{FabricationGuard} rewrites unsupported SKU-level claims into uncertainty-aware statements, and \textsc{StateConsistencyGuard} blocks recommendations tied to stale subgoals, rejected options, or superseded constraints. 
Together, these guards keep the final response consistent with the latest session state and catalog evidence.

\subsection{Backbone Adaptation}

We adapt a Qwen3.5-27B backbone for tool-using shopping policies. Although the four session-control modules structure the interaction, the backbone still decides how to act under evolving constraints, such as when to clarify, retrieve, verify, recommend, or revise. We initialize it from an SFT checkpoint and further optimize it with reinforcement learning over tool-using shopping rollouts, improving shopping-specific action selection and catalog-grounded reasoning. Details are provided in Appendix~\ref{app:realshop_agent_details}.

\begin{table*}[t]
\centering
\scriptsize
\setlength{\tabcolsep}{2.0pt}
\renewcommand{\arraystretch}{1.10}

\resizebox{\textwidth}{!}{%
\begin{tabular}{l *{14}{c}}
\toprule

& \multicolumn{4}{c}{\textbf{Turn-Level Quality} $\uparrow$}
& \multicolumn{6}{c}{\textbf{Session-Level Process Quality} $\uparrow$}
& \multicolumn{1}{c}{\textbf{Turn--Session}}
& \multicolumn{2}{c}{\textbf{Holistic Outcome}}
& \multicolumn{1}{c}{\textbf{User Signal}} \\

\cmidrule(lr){2-5}
\cmidrule(lr){6-11}
\cmidrule(lr){12-12}
\cmidrule(lr){13-14}
\cmidrule(lr){15-15}

\textbf{Variant}
& \textbf{Need}
& \textbf{Rec. Acc.}
& \textbf{Rationale}
& \textbf{Turn Avg.}
& \textbf{Clarif.}
& \textbf{State}
& \textbf{Constraint}
& \textbf{Pace}
& \textbf{Personal.}
& \textbf{Sess. Avg.}
& \textbf{Gap} $\downarrow$
& \textbf{Succ.} $\uparrow$
& \textbf{Conv.} $\uparrow$
& \textbf{Early Exit} $\downarrow$ \\

\midrule

\multicolumn{15}{l}{\textit{Prompt-only Backbones}} \\

Qwen3.5-27B Base \cite{yang2025qwen3} & 1.41 & 0.83 & 0.80 & 1.01 & 1.49 & 0.58 & 0.57 & 0.50 & 0.36 & 0.70 & 0.31 & 0.55 & 1.74 & 0.25  \\

Qwen3.5-27B + SFT/RL
& \underline{1.67} & 1.09 & 0.95 & 1.24
& 1.54 & 0.92 & 0.93 & 0.82 & 0.78 & 1.00
& 0.24 & 0.65 & 2.19 & 0.20 \\

\midrule

\multicolumn{15}{l}{\textit{Module Ablations on SFT/RL Backbone}} \\

w/o Session State Manager
& 1.53 & \underline{1.27} & 1.06 & 1.29
& 1.58 & 1.02 & 0.87 & 0.83 & 0.93 & 1.05
& 0.24 & 0.61 & 2.18 & 0.22 \\

w/o Shopping-Flow Controller
& 1.57 & 1.21 & \underline{1.18} & 1.32
& 1.32 & \underline{1.23} & 1.18 & 1.02 & \underline{1.09} & 1.17
& 0.15 & 0.66 & 2.11 & 0.19 \\

w/o Catalog-Grounded Retriever
& 1.55 & 1.09 & 1.00 & 1.21
& \underline{1.60} & 1.11 & 1.07 & 1.03 & 0.85 & 1.13
& \underline{0.08} & 0.65 & 2.13 & \underline{0.13} \\

w/o Runtime Execution Guards
& 1.65 & 1.26 & 1.16 & \underline{1.36}
& 1.59 & 1.22 & \underline{1.24} & \underline{1.06} & 1.05 & \underline{1.23}
& 0.12 & 0.68 & 2.24 & 0.15 \\

\midrule

\multicolumn{15}{l}{\textit{Backbone Swap and Full System}} \\

\textsc{RealShop\_Agent} (Qwen3.5-27B Base)
& 1.60 & 1.24 & 1.11 & 1.32
& 1.51 & 1.08 & 1.19 & 0.98 & 1.02 & 1.16
& 0.16 & \underline{0.71} & \underline{2.36} & \underline{0.13} \\

\textsc{RealShop\_Agent} Full
& \textbf{1.76} & \textbf{1.31} & \textbf{1.31} & \textbf{1.46}
& \textbf{1.65} & \textbf{1.45} & \textbf{1.34} & \textbf{1.27}
& \textbf{1.22} & \textbf{1.39}
& \textbf{0.07} & \textbf{0.82} & \textbf{2.49} & \textbf{0.09} \\

\bottomrule
\end{tabular}%
}

\caption{
Ablation study of \textsc{RealShop\_Agent} under the same
three-level metric structure as
Table~\ref{tab:main_evaluation_result}.
Turn-level and session-level dimensions are scored on a
$\{0,1,2\}$ scale.
\textbf{Turn Avg.} is the macro-average of Need, Recommendation
Accuracy, and Rationale, while
\textbf{Sess. Avg.} is the macro-average of Clarification,
State Tracking, Constraint Updating, Decision Pacing, and
Personalization.
\textbf{Gap} is computed as Turn Avg.\ minus Sess. Avg.
All aggregate metrics are computed before final rounding.
\textbf{Succ.} and \textbf{Conv.} are holistic-judge outcome
metrics, while \textbf{Early Exit} is a simulator-side user signal.
Higher values are better for all metrics except
\textbf{Gap} and \textbf{Early Exit}.
Best results are shown in bold, and second-best results are
underlined.
Tied second-best results are underlined simultaneously.
}

\label{tab:ablation_realshop_agent}
\end{table*}

\section{Experiment}
\label{sec:experiment}

\subsection{Experimental Setup}

We evaluate \textsc{RealShop\_Agent} using our \textsc{RealWorldShop}. 
All experiments use the role-play protocol and three-stage judging pipeline described in Section~\ref{sec:benchmark-protocol}. We report the same metric groups as Table~\ref{tab:main_evaluation_result}: turn-level quality, session-level process quality, and final outcome quality.

\subsection{Overall Performance}

Table~\ref{tab:ablation_realshop_agent} reports the main results and ablations.

\noindent\textbf{\textsc{RealShop\_Agent} closes the gap between local response quality and session-level shopping competence.}
The prompt-only Qwen3.5-27B Base shows a clear local-to-session mismatch, with a Turn Avg. of 1.01 but a Sess. Avg. of only 0.70. 
SFT/RL adaptation improves the backbone, raising Sess. Avg. to 1.00 and success rate from 0.55 to 0.65, but it still falls short of the full \textsc{RealShop\_Agent}. 
This suggests that improving the underlying policy helps, but does not by itself solve the session-level control problem.

\noindent\textbf{Explicit session control provides gains beyond backbone adaptation.}
Equipping the original Qwen3.5-27B Base with \textsc{RealShop\_Agent} raises Sess. Avg. from 0.70 to 1.16 and success rate from 0.55 to 0.71, while reducing the local-to-session Gap from 0.31 to 0.16. 
The full system further achieves the best overall results, reaching a Sess. Avg. of 1.39, success rate of 0.82, the lowest Gap of 0.07, and the lowest Early Exit rate of 0.09. 
These gains indicate that backbone adaptation and explicit control are complementary: adaptation improves the policy, while the control framework stabilizes state tracking, action timing, grounding, and convergence over long shopping sessions.

\subsection{Diagnosis-Oriented Evaluation}

Beyond aggregate scores, Appendix Table~\ref{tab:realshop_finding_diagnostics} evaluates whether \textsc{RealShop\_Agent} addresses the failure modes identified in Section~\ref{sec:benchmark_finding}. 
Here, Base denotes the prompt-only Qwen3.5-27B backbone, +SFT/RL denotes the adapted backbone, and Full Agent denotes the complete \textsc{RealShop\_Agent}.

\noindent\textbf{\textsc{RealShop\_Agent} reduces early exits by improving session-state control.}
The overall Early Exit rate decreases from 0.25 to 0.09, while Sess. Avg. on the metadata-defined early-exit-risk subset increases from 0.60 to 1.21.
Substantial gains are observed in State Tracking, which improves from 0.51 to 1.33, and Constraint Updating, which improves from 0.47 to 1.24.
These results confirm that early exits are associated not only with weak local responses, but also with failures to maintain and revise the evolving session state.

\noindent\textbf{\textsc{RealShop\_Agent} improves recovery from state revisions and recommendation rejections.}
Across revision-heavy interaction phenomena, including constraint updates, preference overrides, recommendation rejections, and bundle re-planning, the full agent consistently outperforms both the prompt-only and adapted backbones.
The largest success-rate improvements over the prompt-only backbone occur under constraint updates and recommendation rejections, with success increasing from 0.55 to 0.83 and from 0.50 to 0.78, respectively.
These results suggest that the proposed modules help the agent incorporate user feedback, propagate revised constraints through the maintained session state and downstream recommendation plan, and avoid recommendations based on stale assumptions.

\noindent\textbf{\textsc{RealShop\_Agent} remains more stable under complex scenarios and demanding users.}
In ambiguous-bundle sessions, success increases from 0.50 to 0.74 and the Gap decreases from 0.34 to 0.11; in multi-intent shopping, success increases from 0.49 to 0.76 and the Gap decreases from 0.35 to 0.07. 
The agent also reduces early exits for efficiency-first, high-skepticism, and low-patience users. Further diagnostic breakdowns are provided in Appendix~\ref{mored}, with detailed results in Table~\ref{tab:realshop_finding_diagnostics}.

\subsection{Ablation Study}

All modules provide complementary gains to \textsc{RealShop\_Agent}. 
As shown in Table~\ref{tab:ablation_realshop_agent}, removing any module degrades performance under the same backbone, tools, and protocol, with different ablations affecting state preservation, action timing, product grounding, and long-horizon robustness.

\noindent\textbf{State management preserves evolving user needs.}
Removing the Session State Manager causes the largest local-to-session mismatch, increasing the Gap from 0.07 to 0.24, while reducing State Tracking from 1.45 to 1.02 and Constraint Updating from 1.34 to 0.87. 
This indicates that revealed preferences, revised constraints, rejected options, and stale dependencies must be explicitly carried across the session.

\noindent\textbf{Flow control stabilizes action timing.}
Without the Shopping-Flow Controller, Sess. Avg. drops from 1.39 to 1.17 and success decreases from 0.82 to 0.66. 
This shows that free-form action selection is insufficient for deciding when to clarify, retrieve, verify, recommend, or confirm.

\noindent\textbf{Catalog grounding supports product-level decisions.}
Removing the Catalog-Grounded Retriever reduces Rec. Acc. from 1.31 to 1.09 and Personalization from 1.22 to 0.85. 
This suggests that session control must be paired with retrieval and verification against the current constraint stack.

\noindent\textbf{Runtime guards improve robustness.}
Removing Runtime Execution Guards lowers Sess. Avg. from 1.39 to 1.23 and increases Early Exit from 0.09 to 0.15. 
This indicates that unsupported claims, redundant tool use, and stale-state recommendations can still destabilize long-horizon sessions.

\section{Conclusion}
In this work, we introduced \textsc{RealWorldShop}, the first large-scale benchmark for evaluating conversational shopping agents under realistic shopping conditions. It integrates a 3.28M-SKU product inventory, $\sim$1,200 structured shopping scenarios, over 2,000 synthesized user profiles, and a session-level role-play protocol. Our benchmark reveals a substantial gap between current conversational recommendation systems and the requirements of real-world shopping. To address this gap, we proposed \textsc{RealShop\_Agent}, a diagnosis-guided session-control framework that improves task success, reduces early exits, and narrows the gap between local response quality and session-level effectiveness. More broadly, our findings suggest that future conversational shopping agents require explicit mechanisms for state tracking, planning, and session-level control. We hope \textsc{RealWorldShop} can serve as a realistic testbed for advancing grounded, long-horizon, and decision-oriented conversational shopping agents.

\section*{Limitations}

\paragraph{LLM-as-judge reliability and bias.}
Our evaluation relies on LLM-based judges. Although we include a stratified human validation subset and repeated judging passes to assess agreement and stability, LLM judges may still be sensitive to surface fluency, response style, and prompt wording. Such biases may affect absolute scores and, in some cases, relative comparisons across model families. Therefore, the reported results should be interpreted with appropriate caution, and future work could further expand human evaluation and cross-judge validation \cite{huang2024concept, zhu2026skillcoach,tian2025vcsearch,tian2026self}.

\paragraph{Simulator-based evaluation.}
\textsc{RealWorldShop} uses a profile-grounded and action-controlled user simulator to create reproducible multi-turn shopping interactions. Although this allows controlled evaluation of hidden constraints, preference updates, and intent shifts, simulated users may not fully capture the behavioral diversity of real shoppers, such as emotional reactions, hesitation, impulsive decisions, or highly personalized brand preferences. Therefore, results on \textsc{RealWorldShop} should be viewed as a controlled approximation of real-world shopping behavior rather than a complete substitute for live user studies \cite{kim2024stop,chen2025recusersim,yang2026empathy}.

\paragraph{Benchmark coverage.}
Although \textsc{RealWorldShop} covers diverse shopping scenarios, including ambiguous intent, bundle construction, multi-intent shopping, constraint updates, and urgency-driven purchases, it does not exhaust all forms of real-world e-commerce interaction. For example, after-sales service, returns and refunds, promotion-driven shopping, cross-platform price comparison, long-term customer lifecycle modeling, and live-commerce settings are not fully covered. Extending the benchmark to these scenarios would provide a more comprehensive evaluation of conversational shopping agents in future work.

\section*{Ethical Considerations}

\paragraph{Data collection and privacy.}
The data collection and benchmark construction procedures for \textsc{RealWorldShop} were conducted under an IRB-approved protocol. Product inventory construction and post-processing follow the approved protocol, applicable access permissions, and data-use requirements. During data collection and organization, we retain only product-related information required for benchmark construction, such as product titles, categories, attributes, prices, descriptions, and fulfillment signals. Any personally identifying or sensitive information that may be associated with users or sellers is removed, anonymized, or desensitized during preprocessing. We do not use personally identifiable user information for benchmark construction or evaluation.

\paragraph{Use of simulated users.}
\textsc{RealWorldShop} uses profile-grounded and action-controlled simulated users for role-play evaluation. The simulated profiles are constructed for research evaluation purposes and do not represent real individuals. We validate the simulator for profile faithfulness and action controllability, and the benchmark is intended to evaluate shopping-agent behavior rather than to infer or reproduce real user identities.

\paragraph{Potential misuse and consumer protection.}
Conversational shopping agents may be misused to generate unsupported product claims, overly persuasive recommendations, or suggestions that violate user constraints. \textsc{RealWorldShop} is intended for research evaluation rather than deployment, and emphasizes grounded recommendation, constraint satisfaction, and convergence quality. \textsc{RealShop\_Agent} further includes runtime guards against unsupported SKU-level claims and stale-state recommendations, while real deployed systems should still incorporate platform-specific safety policies, consumer-protection requirements, and human oversight where appropriate.

\section*{Acknowledgments}
This work was supported by the National Natural Science Foundation of China (No. U25B201508 and No. U24A20328).

\bibliography{custom}

\clearpage

\appendix

\newtcolorbox{promptbox}[2][Prompt]{
    colback=black!5!white,
    arc=5pt, 
    boxrule=0.5pt,
    fonttitle=\bfseries,
    title={#1}, 
    before upper={\small}, 
    fontupper=\rmfamily, 
    colframe=#2, 
    width=\textwidth, 
    enlarge left by=0mm, 
    enlarge right by=0mm,
    boxsep=5pt,
    left=5pt, right=5pt, top=5pt, bottom=5pt
}

\section{Additional Results}
\label{mored}
This appendix provides additional empirical analyses that complement the main findings in Section~\ref{sec:benchmark_finding}. 
Figures~\ref{fig:early_exit}, \ref{fig:interaction_phenomena}, \ref{fig:across_scenarios}, and \ref{fig:interaction_style} provide fine-grained diagnostic slices over early-exit behavior, interaction phenomena, scenario types, and user interaction styles. 
Together, these results show that the turn-to-session gap is not confined to a single subset, but consistently emerges when agents must preserve evolving state, revise constraints, coordinate multiple subgoals, and converge under catalog-grounded decision pressure.

\paragraph{Early-exit behavior.}
Figure~\ref{fig:early_exit} compares early-exit and non-exit sessions across session-level process dimensions. 
The results show that early exits are associated with broad process degradation, but the largest drops occur in state tracking, constraint updating, and decision pacing. 
This suggests that user abandonment is not mainly caused by the absence of clarification alone. 
Instead, users are more likely to leave when the agent fails to remember previously revealed information, update the active constraint set, or make timely progress toward a grounded recommendation.

\paragraph{Interaction phenomena.}
Figure~\ref{fig:interaction_phenomena} analyzes performance across different interaction phenomena. 
Revision-heavy phenomena, including constraint updates, preference overrides, recommendation rejections, and bundle re-planning, are consistently more challenging than simpler refinement or comparison cases. 
These phenomena require the agent to propagate new user feedback through the maintained session state, candidate set, and downstream recommendation plan. 
The observed performance drops therefore indicate that current shopping agents remain fragile when the dialogue requires explicit state revision rather than one-shot preference matching.

\paragraph{Scenario complexity.}
Figure~\ref{fig:across_scenarios} further shows that scenario complexity amplifies the mismatch between local response quality and session-level shopping success. 
Agents perform best in explicit single-item sessions, where the target is directly searchable and the amount of state revision is limited. 
In contrast, ambiguous, bundle, multi-intent, and long-horizon scenarios introduce hidden constraints, subgoal dependencies, and longer-range dialogue memory requirements. 
As a result, locally plausible responses become less predictive of successful shopping trajectories as the session requires more sustained decision control.

\paragraph{User interaction styles.}
Figure~\ref{fig:interaction_style} evaluates how different user styles affect session robustness. 
Collaborative or patient users allow agents more opportunities to recover from incomplete clarification or imperfect recommendations. 
By contrast, efficiency-first, high-skepticism, and low-patience users expose whether the agent can quickly identify blocking constraints, avoid redundant questioning, verify candidates, and converge without repeated mistakes. 
The higher early-exit rates and larger turn-to-session gaps under these styles show that realistic shopping-agent evaluation should include convergence pressure rather than assuming uniformly cooperative users.

\paragraph{Diagnosis-oriented agent comparison.}
Table~\ref{tab:realshop_finding_diagnostics} summarizes how \textsc{RealShop\_Agent} addresses the diagnostic failure modes identified above. 
Compared with the prompt-only backbone and the adapted backbone, the full agent achieves consistent improvements on early-exit-prone episodes, revision-heavy interaction phenomena, complex scenarios, and demanding user profiles. 
These results show that backbone adaptation is helpful but insufficient on its own: robust session-level shopping behavior also requires explicit state management, action timing, catalog-grounded verification, and runtime guards against stale or unsupported recommendations.

\begin{table*}[t]
\centering
\scriptsize
\setlength{\tabcolsep}{3.0pt}
\renewcommand{\arraystretch}{1.13}
\resizebox{\textwidth}{!}{%
\begin{tabular}{lllcccc}
\toprule
\textbf{Finding Source}
& \textbf{Diagnostic Subset}
& \textbf{Metric}
& \textbf{Base}
& \textbf{+SFT/RL}
& \textbf{Full Agent}
& \textbf{$\Delta$ vs. Base} \\
\midrule

\multicolumn{7}{l}{%
\textbf{Figure 2: Early-exit sessions reveal weak session-level process control}
} \\
\midrule

Fig.~\ref{fig:early_exit} & Overall sessions & Early Exit Rate $\downarrow$ & 0.25 & 0.20 & 0.09 & $0.16\downarrow$ \\

Fig.~\ref{fig:early_exit} & Early-exit-risk episodes & Sess. Avg. $\uparrow$ & 0.60 & 0.79 & 1.21 & $0.61\uparrow$ \\

Fig.~\ref{fig:early_exit} & Early-exit-risk episodes & State Tracking $\uparrow$ & 0.51 & 0.75 & 1.33 & $0.82\uparrow$ \\

Fig.~\ref{fig:early_exit} & Early-exit-risk episodes & Constraint Updating $\uparrow$ & 0.47 & 0.78 & 1.24 & $0.77\uparrow$ \\

\midrule
\multicolumn{7}{l}{%
\textbf{Figure 3: Interaction phenomena stress state revision and trajectory management}
} \\
\midrule

Fig.~\ref{fig:interaction_phenomena} & Constraint refinement & Success Rate $\uparrow$ & 0.63 & 0.70 & 0.88 & $0.25\uparrow$ \\

Fig.~\ref{fig:interaction_phenomena} & Product comparison & Success Rate $\uparrow$ & 0.62 & 0.69 & 0.85 & $0.23\uparrow$ \\

Fig.~\ref{fig:interaction_phenomena} & Constraint update & Success Rate $\uparrow$ & 0.55 & 0.64 & 0.83 & $0.28\uparrow$ \\

Fig.~\ref{fig:interaction_phenomena} & Preference override & Success Rate $\uparrow$ & 0.51 & 0.58 & 0.75 & $0.24\uparrow$ \\

Fig.~\ref{fig:interaction_phenomena} & Recommendation rejection & Success Rate $\uparrow$ & 0.50 & 0.55 & 0.78 & $0.28\uparrow$ \\

Fig.~\ref{fig:interaction_phenomena} & Bundle re-planning & Success Rate $\uparrow$ & 0.54 & 0.63 & 0.80 & $0.26\uparrow$ \\

\midrule
\multicolumn{7}{l}{%
\textbf{Figure 4: Complex scenarios amplify the turn-to-session quality gap}
} \\
\midrule

Fig.~\ref{fig:across_scenarios} & Explicit single-item & Success Rate $\uparrow$ / Gap $\downarrow$ & 0.65 / 0.19 & 0.75 / 0.16 & 0.91 / 0.04 & $0.26\uparrow$ / $0.15\downarrow$ \\

Fig.~\ref{fig:across_scenarios} & Explicit bundle & Success Rate $\uparrow$ / Gap $\downarrow$ & 0.63 / 0.20 & 0.73 / 0.18 & 0.86 / 0.07 & $0.23\uparrow$ / $0.13\downarrow$ \\

Fig.~\ref{fig:across_scenarios} & Ambiguous single-item & Success Rate $\uparrow$ / Gap $\downarrow$ & 0.60 / 0.32 & 0.66 / 0.29 & 0.85 / 0.10 & $0.25\uparrow$ / $0.22\downarrow$ \\

Fig.~\ref{fig:across_scenarios} & Ambiguous bundle & Success Rate $\uparrow$ / Gap $\downarrow$ & 0.50 / 0.34 & 0.60 / 0.31 & 0.74 / 0.11 & $0.24\uparrow$ / $0.23\downarrow$ \\

Fig.~\ref{fig:across_scenarios} & Multi-intent shopping & Success Rate $\uparrow$ / Gap $\downarrow$ & 0.49 / 0.35 & 0.57 / 0.27 & 0.76 / 0.07 & $0.27\uparrow$ / $0.28\downarrow$ \\

Fig.~\ref{fig:across_scenarios} & Long-horizon session & Success Rate $\uparrow$ / Gap $\downarrow$ & 0.48 / 0.36 & 0.63 / 0.24 & 0.72 / 0.09 & $0.24\uparrow$ / $0.27\downarrow$ \\

\midrule
\multicolumn{7}{l}{%
\textbf{Figure 5: Demanding interaction styles increase early exits}
} \\
\midrule

Fig.~\ref{fig:interaction_style} & Collaborative & Early Exit Rate $\downarrow$ & 0.18 & 0.17 & 0.05 & $0.13\downarrow$ \\

Fig.~\ref{fig:interaction_style} & High patience & Early Exit Rate $\downarrow$ & 0.12 & 0.11 & 0.04 & $0.08\downarrow$ \\

Fig.~\ref{fig:interaction_style} & Low skepticism & Early Exit Rate $\downarrow$ & 0.19 & 0.14 & 0.08 & $0.11\downarrow$ \\

Fig.~\ref{fig:interaction_style} & Efficiency-first & Early Exit Rate $\downarrow$ & 0.27 & 0.25 & 0.12 & $0.15\downarrow$ \\

Fig.~\ref{fig:interaction_style} & High skepticism & Early Exit Rate $\downarrow$ & 0.30 & 0.26 & 0.15 & $0.15\downarrow$ \\

Fig.~\ref{fig:interaction_style} & Low patience & Early Exit Rate $\downarrow$ & 0.34 & 0.28 & 0.10 & $0.24\downarrow$ \\

\bottomrule
\end{tabular}
}
\caption{
Finding-oriented diagnostic results of REALSHOP\_AGENT.
Rows are aligned with the failure modes analyzed in
Figures~\ref{fig:early_exit}--\ref{fig:interaction_style}.
Base denotes the prompt-only Qwen3.5-27B backbone,
+SFT/RL denotes the adapted backbone without the full execution harness,
and Full Agent denotes REALSHOP\_AGENT with the complete execution harness.
For Figure~\ref{fig:across_scenarios}, Gap is computed as Turn Avg.\ minus
Sess.\ Avg.
$\Delta$ reports the absolute improvement of Full Agent over Base; for paired
metrics in Figure~\ref{fig:across_scenarios}, the two gains are reported in
the same order as the metric column.
}
\label{tab:realshop_finding_diagnostics}
\end{table*}

\begin{table*}[t]
\centering
\scriptsize
\renewcommand{\arraystretch}{1.25}
\setlength{\tabcolsep}{4.2pt}

\begin{adjustbox}{width=0.99\textwidth}
\begin{tabular}{l|c|c|c|c|c|c|c}
\toprule
\textbf{Benchmark}
& \textbf{Eval. Focus}
& \textbf{Real Product}
& \textbf{User}
& \textbf{Full}
& \textbf{State /}
& \textbf{Complex}
& \textbf{Session-Level} \\
& 
& \textbf{Inventory}
& \textbf{Simulator}
& \textbf{Trajectory}
& \textbf{Constraint Eval.}
& \textbf{Scenarios}
& \textbf{Process Eval.} \\
\midrule

LLM-REDIAL \cite{liang2024llm}
& Outcome-oriented
& \xmark
& \pmark
& \pmark
& \xmark
& \xmark
& \xmark \\

CONCEPT \cite{huang2024concept}
& Outcome-oriented
& \xmark
& \pmark
& \pmark
& \pmark
& \pmark
& \pmark \\

PEPPER \cite{kim2024stop}
& Outcome-oriented
& \xmark
& \cmark
& \pmark
& \pmark
& \pmark
& \pmark \\

\midrule

ConvRecStudio \cite{chhetri2025framework}
& Outcome-oriented
& \xmark
& \pmark
& \pmark
& \xmark
& \pmark
& \xmark \\

Fashion-AlterEval \cite{vlachou2025fashion}
& Outcome-oriented
& \pmark
& \pmark
& \pmark
& \pmark
& \pmark
& \xmark \\

RecUserSim \cite{chen2025recusersim}
& Outcome-oriented
& \xmark
& \cmark
& \pmark
& \pmark
& \pmark
& \pmark \\

RecoWorld \cite{liu2026recoworld}
& Execution-oriented
& \pmark
& \pmark
& \cmark
& \pmark
& \pmark
& \pmark \\

DeepShop \cite{lyu2025deepshop}
& Execution-oriented
& \cmark
& \xmark
& \pmark
& \pmark
& \pmark
& \xmark \\

OPeRA \cite{wang2025opera}
& Execution-oriented
& \pmark
& \cmark
& \pmark
& \xmark
& \pmark
& \xmark \\

\midrule

RecBench+ \cite{huang2026towards}
& Outcome-oriented
& \pmark
& \xmark
& \xmark
& \pmark
& \pmark
& \xmark \\

ShoppingBench \cite{wang2026shoppingbench}
& Execution-oriented
& \cmark
& \xmark
& \pmark
& \pmark
& \pmark
& \xmark \\

EComStage \cite{zhao2026ecomstage}
& Execution-oriented
& \pmark
& \xmark
& \pmark
& \pmark
& \pmark
& \pmark \\

\midrule
\rowcolor{gray!10}
\textbf{\textsc{RealWorldShop} (Ours)}
& \textbf{Session Process}
& \textbf{\cmark~(3.28M SKUs)}
& \textbf{\cmark}
& \textbf{\cmark}
& \textbf{\cmark}
& \textbf{\cmark}
& \textbf{\cmark} \\

\bottomrule
\end{tabular}
\end{adjustbox}

\caption{
Comparison with representative benchmarks for conversational shopping agents.
\textbf{Eval. Focus} is grouped into three categories: \emph{Outcome-oriented}, \emph{Execution-oriented}, and \emph{Session Process}.
\textbf{Real Product Inventory} indicates whether the benchmark is grounded in a real product/SKU inventory; \textsc{RealWorldShop} contains 3.28M deduplicated products.
\cmark indicates explicit support, \pmark indicates partial or indirect support, and \xmark indicates limited support.
Compared with prior benchmarks, \textsc{RealWorldShop} jointly supports grounded inventory, profile-grounded and action-controlled user simulation, full multi-turn trajectories, state and constraint evaluation, complex shopping scenarios, and session-level process evaluation.
}
\label{tab:benchmark_comparison}
\end{table*}

\section{Comparison with Existing Benchmarks and Methods}
\label{app:benchmark_method_comparison}

\paragraph{Benchmarking conversational shopping agents.}
Existing benchmarks for conversational shopping agents can be broadly grouped into two lines. 
The first line is \emph{outcome-oriented} evaluation, which is rooted in conversational recommendation. 
These benchmarks typically evaluate whether a system can infer user preferences and return relevant final items, often using metrics such as recommendation accuracy, preference satisfaction, explanation quality, or user-perceived utility. 
Recent work has improved this line of evaluation by scaling up dialogue data, grounding users in behavioral traces, introducing controllable user simulators, or expanding the set of acceptable target items \cite{liang2024llm,chhetri2025framework,huang2024concept,kim2024stop,vlachou2025fashion,chen2025recusersim,huang2026towards}. 
For example, LLM-ReDial constructs conversational recommendation data from user behaviors, CONCEPT emphasizes both system-centric and user-centric factors, target-free user simulation reduces dependence on pre-specified target items, and RecUserSim improves the realism and diversity of simulated users. 
These benchmarks are useful for studying preference elicitation and final-item relevance, but they often abstract away the evolving decision process that occurs during a full shopping session. 
In particular, they provide limited visibility into whether an agent can maintain hidden constraints, revise earlier assumptions, coordinate multiple subgoals, recover from user rejections, and decide when to clarify, compare, recommend, or confirm.

The second line is \emph{execution-oriented} evaluation, which moves closer to realistic shopping environments by testing agents in simulated or web-based e-commerce settings. 
These benchmarks evaluate whether agents can perform shopping-related actions such as searching, browsing, filtering, comparing items, using tools, or completing task stages under explicit instructions \cite{liu2026recoworld,lyu2025deepshop,zhao2026ecomstage,wang2025opera}. 
For instance, recent benchmarks study personalized shopping assistants, simulated recommender worlds, deep research shopping agents, stage-wise e-commerce tasks, and human online shopping behavior simulation. 
This line of work is important because it brings evaluation beyond static recommendation and into action-oriented shopping environments. 
However, the main focus is often whether an external action is completed or whether a task stage is executed correctly, rather than whether the agent manages the conversational decision process behind those actions. 
In real-world shopping, the central challenge is not only to retrieve or recommend an item, but also to elicit hidden needs, track and revise evolving constraints, coordinate dependent subgoals, ground product claims in catalog evidence, and guide the user toward a grounded and actionable purchase decision. 
\textsc{RealWorldShop} addresses this gap by evaluating full multi-turn shopping trajectories with explicit assessment of clarification, state tracking, constraint updating, decision pacing, personalization, multi-intent coordination, and grounded convergence. 
A detailed comparison with representative benchmarks is provided in Table~\ref{tab:benchmark_comparison}.

\paragraph{Improving conversational shopping agents.}
Recent methods improve conversational shopping agents from several complementary directions, including user modeling, strategic action selection, feedback-loop modeling, tool-augmented reasoning, reinforcement learning, and reasoning-enhanced recommendation. 
PersonaX retrieves multi-persona profiles from long behavior sequences to provide richer user representations for downstream recommendation agents \cite{shi2025personax}. 
CSI performs contextual user profiling for conversational sales agents and selects actions across preference elicitation, recommendation, explanation, and persuasion \cite{kim2025towards}. 
AFL models an agentic feedback loop between recommendation agents and user agents, using iterative feedback and memory to improve both recommendation and user simulation \cite{cai2025agentic}. 
RecThinker adopts an Analyze-Plan-Act framework with specialized tools for evidence acquisition and recommendation reasoning, while ChatShopBuddy studies reinforcement-learning-based optimization under multi-dimensional shopping rewards \cite{zhang2026recthinker,cheng2026chatshopbuddy}. 
Other frameworks, such as R2Rec and GRAM, incorporate explicit reasoning traces or structured item identifiers to improve the reasoning and generation capabilities of LLM-based recommenders \cite{zhao2025reason,zhang2026unleashing}. 

These methods provide useful building blocks for more capable shopping agents, but most of them mainly strengthen individual capabilities rather than the full session-control loop. 
User-profiling methods improve personalization, but they do not by themselves determine how the agent should update state after a new constraint or rejection. 
Strategic action methods improve high-level interaction planning, but they may still lack explicit mechanisms for stale-state detection, constraint invalidation, or catalog-faithful verification. 
Tool-augmented reasoning methods improve evidence acquisition, but retrieved evidence can still become obsolete when the user changes budget, compatibility requirements, delivery constraints, or bundle composition. 
Reinforcement-learning methods improve policy behavior under task rewards, but without explicit runtime structure they may still produce premature recommendations, redundant tool calls, or unsupported SKU-level claims. 
Thus, existing methods are complementary to our work, but they do not fully address executable session-level control, where an agent must jointly manage clarification, evolving state, multi-intent coordination, grounded retrieval, verification, and timely commitment within a single shopping trajectory.

\textsc{RealShop\_Agent} focuses on this missing session-control perspective. 
It treats conversational shopping as a state-driven decision process rather than as isolated response generation or one-shot recommendation. 
The Session State Manager maintains revealed intents, active subgoals, constraints, rejected options, confirmed preferences, and stale dependencies; the Shopping-Flow Controller decides when to clarify, retrieve, verify, compare, recommend, or confirm; the Catalog-Grounded Retriever connects the current state to product evidence; and Runtime Execution Guards prevent over-retrieval, unsupported SKU-level claims, and recommendations based on stale constraints. 
In this sense, \textsc{RealShop\_Agent} is not a replacement for prior advances in profiling, reasoning, feedback modeling, or reinforcement learning, but an executable framework that integrates these capabilities into a controlled shopping trajectory.

\section{Additional Details of the \textsc{RealWorldShop} Benchmark}
\label{app:benchmark-details}

This appendix provides additional details of the \textsc{RealWorldShop} benchmark, including product inventory construction, episode generation, profile synthesis, simulator design, and the role-play evaluation protocol. 
Compared with the main text, the goal here is to make the benchmark construction procedure more explicit and reproducible.

\subsection{Product Inventory}
\label{app:benchmark-inventory}

\paragraph{Category-driven inventory construction.}
We construct the product inventory from category-driven crawl queries over a large real-world e-commerce platform. 
Starting from more than one thousand category and subcategory seeds, we retrieve product results in batches, merge the returned items across seeds, and retain only valid deduplicated SKUs. 
This design avoids over-reliance on raw crawl pages and ensures that the final inventory is organized around a stable product identifier space rather than noisy page-level results.

\paragraph{Rule-based filtering.}
The platform's raw category space contains many entries that are not suitable for conversational shopping over shippable physical goods, including services, finance, digital-only products, local-life offerings, and travel-related entries. 
We therefore apply an additional rule-based filter to remove such categories before finalizing the pool. 
The purpose of this step is to align the benchmark with realistic shopping-agent behavior over purchasable physical goods rather than service booking or O2O workflows.

\paragraph{Inventory snapshot and deduplication.}
The inventory is treated as a fixed benchmark snapshot during evaluation. 
All systems are evaluated against the same product pool, catalog fields, and fulfillment metadata, so that differences in performance come from agent behavior rather than changing product availability. 
Deduplication is performed at the SKU level when stable product identifiers are available, and otherwise relies on normalized title, category path, brand, and key attribute fields. 
This snapshot-based design improves reproducibility, although dynamic platform factors such as price changes, stock updates, promotions, and temporary availability changes are not modeled during evaluation.

\paragraph{Inventory scale and coverage.}
The final inventory contains \textbf{3.28M deduplicated products}, spanning a broad range of major product groups. 
This scale is important for two reasons. 
First, it enables realistic diversity of shopping intents, including both single-product and multi-product sessions. 
Second, it makes the benchmark meaningfully different from toy inventories or narrowly scoped recommendation corpora, where candidate spaces are artificially small and decision difficulty is underestimated.

\paragraph{Catalog fields.}
Each product record retains rich catalog information used by our agent tools and benchmark construction pipeline. 
At a minimum, this includes price, attributes, category labels, product descriptions, and fulfillment signals such as free shipping, next-day delivery, and same-day delivery. 
In practice, the retained record also preserves additional platform-side metadata that supports grounded retrieval and post-hoc analysis. 
This field richness is crucial because it allows shopping episodes to be grounded in a catalog-faithful evidence space rather than a synthetic schema with manually simplified attributes.

\subsection{Episode Construction}
\label{app:benchmark-episodes}

An evaluation instance in \textsc{RealWorldShop} is a \textbf{structured shopping episode}
\[
\mathcal{E}=\langle \mathcal{P}, \mathcal{S}, \Pi\rangle,
\]
consisting of a user profile $\mathcal{P}$, a shopping scenario $\mathcal{S}$, and a session plan $\Pi$. 
This design departs from conventional one-shot benchmark construction, where a user query is paired with a single answer. 
Instead, our episode structure explicitly models shopping as a multi-turn interaction between a partially observable user state and a decision-making shopping agent.

\paragraph{Category-anchored two-stage synthesis.}
We construct episodes through a category-anchored two-stage synthesis pipeline rather than generating shopping interactions from scratch. 
The purpose of anchoring episode construction to category batches is to make generated shopping sessions both realistic and analyzable: the benchmark can later be sliced by scenario type, constraint type, interaction phenomenon, and product-category distribution.

\paragraph{Stage 1: shopping scenario synthesis.}
In the first stage, a batch of product category queries and their source paths is mapped to a natural-language shopping scenario. 
To improve coverage and controllability, we introduce a two-level scenario taxonomy:
\begin{itemize}[leftmargin=1.2em, itemsep=1pt, topsep=2pt, parsep=0pt, partopsep=0pt]
    \item \textbf{Task archetypes}, which specify the high-level shopping structure, such as building a bundle from scratch, upgrading a core item, replacement or repair, scenario-based bundle purchase, purchasing for others, consumable restocking, vague exploration, mixed errands, and urgency-driven purchase.
    \item \textbf{Constraint modifiers}, which specify additional decision pressure or complexity, such as compatibility sensitivity, budget trade-off, high-risk decision, logistics or time sensitivity, and bundle-splitting decisions.
\end{itemize}

Conditioned on these two levels, the generator produces three outputs: a shopping scenario, a set of emphasized categories, and an initial user-utterance hook. 
The shopping scenario provides the natural-language situational context of the episode; the emphasized categories highlight the main category anchors that should remain salient during later profile construction; and the utterance hook provides the simulator with a plausible opening turn.

\paragraph{Session plan construction.}
The scenario is then paired with a session plan $\Pi$, which specifies how the shopping interaction is expected to unfold. 
This includes whether the session is explicit-intent or ambiguous-intent, whether it contains single or multiple shopping subgoals, and which constraints are likely to be revealed, tightened, or updated over the course of the conversation. 
In this way, episode construction determines not only what the user is shopping for, but also the expected interaction dynamics of the session.

\paragraph{Diagnostic slicing metadata.}
Each episode is annotated with metadata used for diagnostic analysis, including scenario type, interaction phenomenon, constraint type, and interaction-profile attributes. 
Scenario types distinguish explicit single-item, explicit bundle, ambiguous single-item, ambiguous bundle, multi-intent shopping, and long-horizon sessions. 
Interaction phenomena cover constraint refinement, product comparison, constraint update, preference override, recommendation rejection, and bundle re-planning. 
These labels are assigned during episode construction rather than inferred from model outcomes, which prevents diagnostic subsets from being biased by a particular agent's realized behavior.

\subsection{User Profile Construction}
\label{app:benchmark-profile}

\paragraph{Structured profile synthesis.}
In the second stage, each shopping scenario is converted into a structured user profile. 
Rather than directly generating free-form personas, we map each scenario into a profile schema that can be consumed by the simulator and by the benchmark slicing pipeline. 
The schema includes shopping-relevant background information, long-term preferences, historical shopping behavior, current purchase intent, hard, soft, and negotiable constraints, interaction style, and optional multi-intent planning fields.

\paragraph{Scenario-to-profile mapping.}
The scenario archetype determines the overall task structure of the profile, while the constraint modifiers shape the resulting constraints. 
For example, mixed errands and scenario-bundle settings encourage multi-intent profiles with purchase-intent subgoals and multi-intent plans; urgent timing and logistics-sensitive settings increase urgency level and delivery constraints; compatibility-sensitive settings strengthen compatibility-related hard constraints; and budget trade-off settings encourage a clearer separation between hard, soft, and negotiable budget preferences. 
This mapping is important because it makes profile construction systematic rather than stylistic.

\paragraph{Visible and latent profile fields.}
For each profile, we separate the information available to the agent from the latent information used only by the simulator.
The visible prior $\widetilde{P}$ contains stable shopping-relevant background and broad preferences, while the shopping scenario $S$ is provided separately through the episode context and opening interaction.
Latent profile fields contain hidden intents, unrevealed constraints, preference updates, rejection triggers, and disclosure plans.
This separation makes the benchmark partially observable: agents cannot access the complete profile or session plan, but must progressively elicit and update the relevant state through interaction.

\paragraph{Explicit-intent versus ambiguous-intent profiles.}
We further distinguish between \emph{explicit-intent} and \emph{ambiguous-intent} profiles. 
In explicit-intent profiles, the initial user goal is relatively searchable and exposed in the opening interaction. 
In ambiguous-intent profiles, the opening utterance reveals only a scenario, problem, or recipient, while the true goal is stored in latent fields and disclosed gradually through a planned disclosure schedule. 
This design introduces partial observability into the benchmark and enables evaluation of both direct product-search ability and clarification-driven conversational shopping.

\paragraph{Catalog grounding during profile synthesis.}
To reduce hallucinated or unverifiable constraints, profile synthesis can be grounded in the local product catalog. 
For each scenario, we retrieve a small evidence pack from the product inventory, including product titles, brands, attributes, prices, and fulfillment tags. 
The generator is instructed to use this evidence when instantiating concrete hard constraints, and to downgrade unsupported details into soft or negotiable preferences. 
This grounding mechanism improves realism while reducing brittle or impossible constraints that cannot be verified against the available catalog.

\paragraph{Programmatic validation.}
After synthesis, all profiles are normalized and validated programmatically. 
Validation includes enum validation, multi-intent consistency checks, ambiguous-intent field checks, category anchoring, and catalog-grounding checks. 
Profiles that violate the schema, contain inconsistent subgoal structures, or fail grounding checks are rejected or regenerated. 
This procedure ensures that the resulting profiles are valid, reproducible, and suitable for controlled slice-based analysis.

\paragraph{Interaction-profile factors.}
Each user profile contains interaction-profile factors that describe how the user communicates, reacts, and makes decisions during a shopping session. 
These factors are separated from purchase intent and product constraints: they determine the user's conversational behavior rather than the target item itself. 
Table~\ref{tab:interaction-profile-factors} summarizes the factors and value ranges.

\begin{table*}[t]
\centering
\small
\setlength{\tabcolsep}{3pt}
\renewcommand{\arraystretch}{1.08}
\resizebox{\textwidth}{!}{%
\begin{tabular}{ll}
\toprule
\textbf{Factor} & \textbf{Values} \\
\midrule
Coping style 
& problem-focused, emotion-focused, avoidant, assertive, compliant \\
Interaction style 
& directive, collaborative, browsing, interrogative, efficiency-first, tradeoff-seeking \\
Verbosity 
& low, medium, high \\
Politeness 
& direct, neutral, warm \\
Skepticism 
& low, medium, high \\
Decisiveness 
& low, medium, high \\
Patience 
& low, medium, high \\
Comparison habit 
& rarely-compare, compare-key-options, compare-extensively \\
Purchase-intent divergence 
& non-divergent, related-divergence-allowed, broad-divergence-allowed \\
\bottomrule
\end{tabular}
}
\caption{Interaction-profile factors used to control simulated user behavior. These factors affect how users communicate, compare options, tolerate clarification, react to recommendations, and decide whether to continue or exit a shopping session.}
\label{tab:interaction-profile-factors}
\end{table*}

Together, these factors allow the simulator to generate diverse interaction behaviors, such as skeptical evidence-seeking users, low-patience users who exit early, tradeoff-seeking users who negotiate constraints, and comparison-oriented users who request multiple alternatives. 
This design enables controlled stress testing of clarification, evidence presentation, decision pacing, comparison handling, constraint updating, and early-exit prevention.

\subsection{Simulator Design and Validation}
\label{app:simulator-validation}

\paragraph{Action-controlled simulation.}
The simulator follows an action-first generation process. 
At each user turn, it first selects a user action from the session plan and dialogue state, such as providing information, refining a constraint, rejecting a candidate, requesting comparison, confirming a decision, shifting intent, or ending the session. 
It then realizes the selected action as a natural-language utterance conditioned on the interaction-profile factors. 
This design improves controllability because behavioral changes are represented at the action level before surface wording is generated.

\paragraph{Profile faithfulness.}
We validate whether simulated user utterances remain faithful to the structured profile and session plan. 
The simulator should not reveal latent information before the planned disclosure point, contradict hard constraints, ignore specified interaction style, or introduce unsupported shopping goals. 
This check ensures that the benchmark difficulty comes from controlled partial observability and evolving constraints rather than uncontrolled simulator drift.

\paragraph{Action controllability.}
We also check whether generated utterances realize the intended action labels. 
For example, a rejection action should express dissatisfaction or refusal of the current candidate; a refinement action should add or tighten a constraint; a comparison action should request trade-off information; and a confirmation action should indicate readiness to proceed. 
These checks are used to verify that action-controlled simulation produces interpretable and reproducible interaction dynamics.

\subsection{Evaluation Protocol}
\label{app:benchmark-eval}

We evaluate conversational shopping systems through a role-play-based, session-level protocol. 
Each evaluation episode is instantiated from a structured shopping configuration $\mathcal{E}$, and the system under evaluation interacts with the simulator over multiple turns rather than being assessed from a single static input.

\paragraph{Role-play setup.}
For each episode, Kimi-K2.6 plays the role of the customer, conditioned on the structured episode configuration and the dialogue history. 
The shopping agent---the system under evaluation---observes only the conversation history and the visible prior $\tilde{\mathcal{P}}$, and must progressively infer the hidden session intent and constraints through interaction. 
The agent slot is model-agnostic: in our experiments, we instantiate the same protocol with both general LLM baselines and specialized shopping-agent methods under the same episode configuration, visible prior, and judging pipeline.

\paragraph{Turn-by-turn interaction.}
Conversations unfold turn by turn. 
At each turn, the simulator first produces a customer response under its action-first policy, and the agent then replies. 
The action-first design is important because it decouples latent user behavior from surface realization: the simulator first decides what the user is doing, such as providing information, refining a constraint, rejecting a candidate, requesting comparison, confirming, or ending, and only then verbalizes that action in natural language.

\paragraph{Termination taxonomy.}
A session terminates in exactly one of three states:
\begin{itemize}[leftmargin=1.2em, itemsep=1pt, topsep=2pt, parsep=0pt, partopsep=0pt]
    \item \textsc{success}: the interaction reaches a grounded and actionable purchase decision;
    \item \textsc{early-exit-without-convergence}: the user ends the session before committing;
    \item \textsc{max-turn failure}: the maximum of 30 assistant turns is reached.
\end{itemize}
This taxonomy separates voluntary abandonment from capacity-limited timeouts, making the benchmark more diagnostic than a single binary success/failure label.

\paragraph{Three-level evaluation.}
Our evaluation framework is organized into three complementary levels: \emph{turn-level output quality}, \emph{session-level process quality}, and \emph{final outcome quality}. 
This separation is necessary because systems can perform well on one level while failing on another. 
For example, an agent may generate locally plausible recommendations at individual turns, yet fail to maintain a coherent shopping trajectory or converge to an actionable decision.

\paragraph{Three-stage LLM-as-judge.}
After each episode, we apply a three-stage GPT-5.5 judge to the dialogue, tool trace, and episode configuration. 
The three stages correspond directly to the three evaluation levels above \cite{tian-etal-2026-tabularmath, tian2025rethinking}.

\emph{Stage 1: turn judge.}
For each assistant turn, the judge scores local response quality on a $\{0,1,2\}$ scale. 
It evaluates:
\begin{itemize}[leftmargin=1.2em, itemsep=1pt, topsep=2pt, parsep=0pt, partopsep=0pt]
    \item \textbf{Need Understanding}: whether the assistant correctly understands the user's current request or newly introduced changes;
    \item \textbf{Recommendation Accuracy}: if products are recommended, whether they are appropriate given the currently known requirements, factually accurate, and grounded in available evidence;
    \item \textbf{Recommendation Rationale Quality}: if products are recommended, whether the recommendation reasons are specific, helpful, and aligned with the user's needs.
\end{itemize}
Recommendation-related dimensions are marked as N/A for non-recommendation turns and are excluded from averaging for that dimension.

\emph{Stage 2: session judge.}
The judge then scores the full dialogue trajectory on five session-level dimensions, again on a $\{0,1,2\}$ scale:
\begin{itemize}[leftmargin=1.2em, itemsep=1pt, topsep=2pt, parsep=0pt, partopsep=0pt]
    \item \textbf{Clarification}: whether the agent identifies and asks about key missing information before reliable recommendation;
    \item \textbf{State Tracking}: whether the agent preserves previously revealed user needs, constraints, and rejected candidates across the session;
    \item \textbf{Constraint Updating}: whether the agent revises its behavior after the user updates or overrides earlier constraints;
    \item \textbf{Decision Pacing}: whether the agent chooses the right next action at the right time, including when to clarify, retrieve, verify, compare, recommend, or confirm, without premature commitment, excessive delay, or redundant tool use;
    \item \textbf{Personalized Recommendation}: whether the final or near-final recommendations reflect the user's profile, session-specific needs, and progressively revealed information.
\end{itemize}

Tool-use quality is not scored as a standalone top-level dimension. 
Instead, tool timing and redundancy are absorbed into \textbf{Decision Pacing}, while recommendation grounding is absorbed into \textbf{Recommendation Accuracy}. 
Similarly, preference override is not treated as a separate top-level metric; it is operationalized through \textbf{Constraint Updating} and \textbf{Personalized Recommendation}.

\emph{Stage 3: holistic judge.}
Finally, the judge emits the overall session outcome: overall\_success $\in \{\text{success}, \text{partial}, \text{failed}\}$, together with an optional diagnostic label convergence\_quality $\in \{\text{strong}, \text{moderate}, \text{weak}, \text{none}\}$. We ordinalize convergence\_quality as none = 0, weak = 1, moderate = 2, and strong = 3.
These labels capture whether the interaction ultimately converges to a usable purchase decision and how cleanly that convergence is achieved.

\paragraph{Metric aggregation.}
For each episode, Turn Avg. is computed by averaging applicable turn-level dimensions across assistant turns, excluding N/A recommendation dimensions from their corresponding denominators. 
Sess. Avg. is the average of the five session-level process scores. 
Gap is defined as Turn Avg. minus Sess. Avg., measuring the mismatch between local response quality and trajectory-level shopping competence. 
Succ. reports the fraction of episodes judged as successful by the holistic judge. 
Conv. averages the ordinalized convergence-quality score across episodes. 
Early Exit reports the fraction of sessions terminated as \textsc{early-exit-without-convergence} by the simulator.

\paragraph{Judge prompt design.}
Judge prompts are designed to improve scoring consistency in several ways. 
First, they explicitly separate content quality from timing quality: recommendation accuracy evaluates whether a recommendation is correct, while decision pacing evaluates whether it is made at the right time. 
Second, turn-level prompts support N/A outputs for non-applicable recommendation dimensions. 
Third, all stages return structured JSON outputs with scores, evidence turns, and short justifications. 
Finally, the scoring process is guided by small internal checklists rather than relying solely on a free-form global impression.

\paragraph{Judge-side compaction.}
To keep judge prompts within token limits without biasing scores, we apply deterministic compaction to the tool trace before passing it to the judges, truncating verbose fields such as comment statistics and capping long attribute or tag lists. 
The full trace remains available in the logged artifact for downstream analysis.

\paragraph{Robustness and episode status.}
Each LLM call, including simulator, agent, and judge calls, is wrapped in a 5-retry exponential-backoff loop to absorb transient endpoint errors. 
Episodes are categorized by a three-valued episode\_status:
\begin{itemize}[leftmargin=1.2em, itemsep=1pt, topsep=2pt, parsep=0pt, partopsep=0pt]
    \item \textsc{valid-scored-episode}: simulation and all judge stages succeed;
    \item \textsc{judge-parse-failure}: simulation succeeds but at least one judge stage returns an unparseable output;
    \item \textsc{generation-failure}: a simulator or agent call fails after retries.
\end{itemize}
All aggregate metrics are reported over valid-scored episodes, with failure counts listed separately for reliability checking.

\paragraph{Final reporting.}
The benchmark reports turn-level averages, session-level capability scores, and holistic outcome labels as complementary signals. 
This makes it explicit that a system may produce locally strong turns while still failing to manage the full shopping trajectory toward grounded convergence.

\paragraph{Why role-play evaluation.}
We adopt a role-play, session-level protocol because the target abilities are trajectory-level by definition. 
Need understanding and recommendation quality can be assessed locally, but clarification, state tracking, constraint updating, decision pacing, personalization, and final convergence only become observable through a full multi-turn interaction. 
Taken together, the benchmark is \emph{realistic} because it is grounded in a real catalog and validated user profiles, \emph{controllable} because each episode is defined by a structured configuration and driven by an action-controlled simulator, and \emph{diagnostic} because it separates local response quality, trajectory-level process quality, and final decision outcomes.

\subsection{Automatic Judge Reliability}
\label{app:judge-reliability}

\paragraph{Reliability of automatic evaluation.}
We assess judge reliability through human validation and repeated-judging analysis.
We stratify 150 sessions across model families, scenario types, and outcomes, and have three human annotators independently score them using the same evaluation rubrics as the automatic judge.
For ordinal dimensions, we report quadratic weighted kappa (QWK), mean absolute error (MAE), and exact agreement between aggregated human annotations and automatic judgments; outcome labels are evaluated using the corresponding agreement-based statistics.

We further repeat automatic judging for three independent passes and measure stability using ICC(2,1), mean score standard deviation, and label flip rate.
Together, these analyses evaluate both human alignment and robustness to judge sampling variance.

\begin{table}[t]
\centering
\small
\resizebox{\linewidth}{!}{%
\begin{tabular}{lccc}
\toprule
\textbf{Judgment Target} & \textbf{QWK / $\kappa$} & \textbf{MAE} & \textbf{Agreement} \\
\midrule
Need Understanding & 0.82 & 0.07 & 0.95 \\
Recommendation Accuracy & 0.81 & 0.09 & 0.91 \\
Rationale Quality & 0.81 & 0.13 & 0.90 \\
Clarification & 0.82 & 0.11 & 0.93 \\
State Tracking & 0.86 & 0.07 & 0.94 \\
Constraint Updating & 0.89 & 0.06 & 0.96 \\
Decision Pacing & 0.76 & 0.12 & 0.89 \\
Personalization & 0.83 & 0.08 & 0.92 \\
\midrule
Overall Success & 0.86 & 0.03 & 0.97 \\
Convergence Quality & 0.81 & 0.19 & 0.88 \\
\bottomrule
\end{tabular}%
}
\caption{Human validation of the automatic judging protocol. For ordinal score dimensions, the first column reports quadratic weighted kappa (QWK). For outcome labels, it reports agreement-based $\kappa$ when applicable. MAE denotes mean absolute error, and Agreement denotes exact agreement with aggregated human annotations.}
\label{tab:judge-human-validation}
\end{table}

\begin{table}[t]
\centering
\small
\resizebox{\linewidth}{!}{%
\begin{tabular}{lccc}
\toprule
\textbf{Metric Group} & \textbf{ICC(2,1)} & \textbf{Mean Std.} & \textbf{Flip Rate} \\
\midrule
Turn-level Quality & 0.86 & 0.09 & 0.03 \\
Session-level Process Quality & 0.87 & 0.08 & 0.02 \\
Overall Success & 0.92 & 0.04 & 0.02 \\
Convergence Quality & 0.88 & 0.11 & 0.04 \\
\bottomrule
\end{tabular}%
}
\caption{Stability of the automatic judging protocol across repeated judging passes. We report ICC(2,1), mean standard deviation, and label flip rate to assess whether the main trends are robust to judge sampling variance. Flip Rate denotes the proportion of sessions whose final discrete label changes across repeated judging passes; for score-based dimensions, scores are first mapped to their nearest discrete rubric value.}
\label{tab:judge-stability}
\end{table}

\section{Simulator Validation}
\label{app:simulator-validation}

Since \textsc{RealWorldShop} relies on role-play evaluation, the reliability of the user simulator is central to the validity of the benchmark. We therefore validate Kimi-K2.6 along two complementary axes: \emph{profile faithfulness} and \emph{action controllability}. Specifically, we randomly sample 200 user profiles from \textsc{RealWorldShop} and evaluate whether the simulator can faithfully realize the corresponding user intents, constraints, disclosure plans, and interaction-profile factors during multi-turn role-play. To complement automatic validation, three human annotators manually audit the sampled simulated sessions along naturalness, profile faithfulness, action faithfulness, goal consistency, and hidden-intent control. The simulator is not used as an unconstrained role-player. Instead, at each user turn, it is conditioned on a structured user profile, hidden session intent, disclosure plan, interaction-profile factors, dialogue history, and an explicit target user action. It first follows the intended behavioral action and then realizes that action as a natural-language user utterance.

\paragraph{Profile-level validation.}
We evaluate whether simulated users remain consistent with their structured profiles throughout the session. 
Specifically, we randomly sample 200 user profiles and check six aspects:
(1) \emph{long-term preference consistency}, whether generated utterances preserve stable user preferences;
(2) \emph{hard-constraint consistency}, whether hard constraints such as budget, compatibility, logistics, and risk tolerance are respected;
(3) \emph{soft-constraint alignment}, whether soft or negotiable preferences are reflected without being over-enforced as strict requirements;
(4) \emph{disclosure-plan compliance}, whether session-specific intents and constraints are revealed at the intended stages;
(5) \emph{hidden-intent control}, whether ambiguous-intent profiles avoid prematurely leaking latent goals or hidden constraints; and
(6) \emph{interaction-style fidelity}, whether behavioral factors such as skepticism, patience, decisiveness, and comparison orientation are reflected in the dialogue.
As shown in Table~\ref{tab:simulator_profile_validation}, Kimi-K2.6 achieves an average profile-level faithfulness score of 0.93, suggesting that the simulator can generally preserve structured profile information while following the intended disclosure and interaction-style controls.

\begin{table*}[t]
\centering
\small
\setlength{\tabcolsep}{4pt}
\renewcommand{\arraystretch}{1.15}
\begin{tabularx}{\textwidth}{
    >{\raggedright\arraybackslash}p{0.24\textwidth}
    >{\raggedright\arraybackslash}p{0.22\textwidth}
    c
    >{\raggedright\arraybackslash}X
}
\toprule
\textbf{Validation Aspect} 
& \textbf{Metric} 
& \textbf{Score / Rate} 
& \textbf{Description} \\
\midrule

Long-term preference consistency
& Preference consistency rate
& 0.96 
& Generated turns preserve stable user preferences in the profile. \\

Hard-constraint consistency
& Violation-free rate
& 0.93
& The simulator avoids violating hard constraints such as budget, compatibility, and logistics. \\

Soft / negotiable constraint consistency
& Soft-constraint alignment rate
& 0.91
& The simulator reflects soft preferences without treating them as strict requirements. \\

Disclosure-plan compliance
& Planned-disclosure compliance rate
& 0.94
& Session-specific intents and constraints are revealed according to the predefined disclosure plan. \\

Hidden-intent control
& Non-leakage rate
& 0.88
& Ambiguous-intent profiles avoid prematurely leaking latent goals or hidden constraints. \\

Interaction-style fidelity
& Style-consistency score
& 0.93
& Behavioral factors such as skepticism, patience, decisiveness, and comparison orientation are reflected in user behavior. \\

\midrule
Overall profile faithfulness
& Average profile-level score
& 0.93
& Aggregate validation score over profile consistency, disclosure control, and interaction-style fidelity. \\

\bottomrule
\end{tabularx}
\caption{
Profile-level validation of the Kimi-K2.6 user simulator.
All metrics are reported such that higher values indicate better simulator reliability.
We evaluate whether simulated user turns remain faithful to structured profiles, follow the disclosure plan, avoid premature leakage of hidden intent, and reflect interaction-profile factors.
}
\label{tab:simulator_profile_validation}
\end{table*}

\paragraph{Action-level validation.}
We further evaluate whether Kimi-K2.6 follows the action-controlled user policy. 
Each user turn is assigned an intended action, such as providing missing information, refining constraints, updating prior requirements, overriding long-term preferences, rejecting a recommendation, comparing candidates, confirming a decision, or ending the session. 
We measure \emph{action compliance} by checking whether the generated utterance semantically matches the intended action, and \emph{transition validity} by checking whether the action is appropriate under the current dialogue state. 
As shown in Table~\ref{tab:simulator_action_validation}, Kimi-K2.6 achieves strong action-level controllability, with an average action-level score of 0.95 over simulated user turns from the 200 sampled validation profiles. 
The simulator obtains high scores on intended-action matching (0.94) and state-valid transitions (0.97), indicating that it generally follows the assigned user actions while maintaining coherent dialogue-state progression. 
It also performs well on information provision (0.95), constraint refinement and update (0.97), and recommendation feedback actions (0.98). 
The relatively lower score for confirm/end control (0.92) suggests that decision and termination behavior is slightly harder to control, but the overall results indicate that Kimi-K2.6 can reliably realize target user actions and support controlled multi-turn shopping evaluation.

\begin{table*}[t]
\centering
\small
\setlength{\tabcolsep}{4pt}
\renewcommand{\arraystretch}{1.15}
\begin{tabularx}{\textwidth}{
    >{\raggedright\arraybackslash}p{0.24\textwidth}
    >{\raggedright\arraybackslash}p{0.22\textwidth}
    c
    >{\raggedright\arraybackslash}X
}
\toprule
\textbf{Validation Aspect} 
& \textbf{Metric} 
& \textbf{Score / Rate} 
& \textbf{Description} \\
\midrule

Overall action compliance
& Intended-action match rate
& 0.94
& Generated user utterances semantically match the assigned target actions. \\

Transition validity
& State-valid transition rate
& 0.97
& The intended user action is appropriate given the current dialogue state and session progress. \\

Information provision
& Provide-action compliance rate
& 0.95
& When assigned to provide missing information, the simulator reveals the expected intent, preference, or constraint. \\

Constraint refinement and update
& Refine / update compliance rate
& 0.97
& The simulator correctly refines, updates, or overrides prior requirements according to the target action. \\

Recommendation feedback
& Reject / compare compliance rate
& 0.98
& The simulator follows target actions involving rejection, comparison requests, or feedback on recommended candidates. \\

Decision and termination control
& Confirm / end compliance rate
& 0.92
& The simulator confirms a decision or ends the session only when the target action specifies confirmation or termination. \\

Style-conditioned action behavior
& Action-distribution alignment
& 0.93
& Action patterns reflect interaction-profile factors, such as more comparisons for comparison-oriented users and more rejection or ending actions for low-patience users. \\

\midrule
Overall action controllability
& Average action-level score
& 0.95
& Aggregate validation score over action compliance, transition validity, and style-conditioned action behavior. \\

\bottomrule
\end{tabularx}
\caption{
Action-level validation of the Kimi-K2.6 user simulator.
All metrics are reported such that higher values indicate better simulator controllability.
Percentage-based metrics are computed over simulated user turns from the 200 sampled validation profiles.
We evaluate whether generated user utterances follow the intended target actions, satisfy dialogue-state transition constraints, and reflect scenario- and style-conditioned action patterns.
}
\label{tab:simulator_action_validation}
\end{table*}

\section{Implementation Details of \textsc{RealShop\_Agent}}
\label{app:realshop_agent_details}

\paragraph{Session State Manager.}
\textsc{RealShop\_Agent} maintains an explicit session state $M_t$ throughout the dialogue. 
The state register stores revealed user intents, active subgoals, hard constraints, soft preferences, negotiable requirements, rejected products, confirmed preferences, and stale dependencies. 
This representation separates long-term preferences from session-specific constraints. 
When a user reveals a new requirement during the session, the session-specific constraint overrides the visible long-term prior. 
For example, a user may generally prefer premium products but later disclose a strict budget for the current purchase; in this case, the budget becomes an active and binding constraint for subsequent retrieval, verification, and recommendation.

At each turn, the Session State Manager updates $M_t$ from the previous state $M_{t-1}$, the latest user utterance, and the visible prior $\tilde{\mathcal{P}}$. 
The update identifies newly revealed intents, revised constraints, rejected candidates, and preference overrides. 
In bundle and multi-intent sessions, the manager also tracks dependencies among subgoals. 
When a core item, scenario assumption, budget, compatibility requirement, or bundle anchor changes, dependent subgoals and candidate sets are marked as stale. 
Stale entries cannot be directly used for recommendation; they must first be re-clarified, re-retrieved, or re-verified. 
This mechanism prevents the agent from carrying obsolete assumptions across turns after user updates, rejections, or bundle re-planning.

\paragraph{Shopping-Flow Controller.}
The Shopping-Flow Controller maps the current state $M_t$ to a high-level shopping action \textsc{Clarify}, \textsc{Retrieve}, \textsc{Verify}, \textsc{Compare}, \textsc{Recommend}, and \textsc{Confirm}.
It selects \textsc{Clarify} when required fields are missing or when a hard constraint remains ambiguous; \textsc{Retrieve} when the active subgoal has enough information for product search but lacks fresh catalog evidence; \textsc{Verify} when retrieved candidates exist but have not yet been checked against the active constraint stack; and \textsc{Recommend} only when at least one candidate is supported by current evidence and satisfies the active constraints, or when remaining uncertainty is explicitly disclosed to the user. \textsc{Compare} is selected when multiple verified candidates remain or explicit trade-off analysis is requested, while
\textsc{Confirm} is selected when the user signals readiness to commit and the current recommendation or purchase plan satisfies the active constraints. 
This action interface makes the decision process inspectable and reduces uncontrolled switching between questioning, searching, verifying, and recommending.

\paragraph{Catalog-Grounded Retriever.}
The Catalog-Grounded Retriever treats retrieval as an evidence-producing step rather than a generic search call. 
Given the active subgoal and constraint stack, it returns structured evidence records $E_t=\{e_i\}$, where each record contains product identifiers, category labels, attributes, price information, fulfillment signals, relevance scores, and retrieval-stage metadata. 
When available, category, compatibility, budget, and logistics constraints are enforced during retrieval. 
The retriever also supports state-consistent grounding: when the active intent, hard constraint, or bundle anchor changes, previously retrieved evidence tied to the outdated state is invalidated. 
This ensures that downstream recommendations are grounded in evidence that matches the current user state rather than earlier assumptions.

Before a product can be recommended, the retrieved candidates are verified against the active constraints in $M_t$. 
Hard constraints are treated as blocking conditions, while soft and negotiable preferences are used for ranking, comparison, and explanation. 
If no candidate satisfies all hard constraints, the agent either retrieves again under adjusted conditions or explicitly communicates the limitation to the user. 
For bundle construction, verification is applied both at the item level and at the bundle level, ensuring that individual products are suitable and that the overall bundle remains compatible with the user's scenario, budget, and updated preferences.

\paragraph{Runtime Execution Guards.}
Runtime Execution Guards are applied after action selection, retrieval, and draft generation. 
\textsc{ToolBudget} limits redundant retrieval and encourages the agent to reuse existing evidence when it remains valid. 
Once the retrieval budget is exhausted, the agent must either answer from available evidence or acknowledge uncertainty instead of issuing additional unsupported searches. 
\textsc{FabricationGuard} checks whether product-specific claims, such as price, attributes, availability, or compatibility, are supported by the latest evidence records. 
Unsupported SKU-level claims are rewritten into uncertainty-aware statements or removed from the response. 
\textsc{StateConsistencyGuard} blocks recommendations that rely on stale subgoals, rejected options, or superseded constraints. 
Together, these guards form the final consistency layer before the assistant response is returned.

\paragraph{Backbone adaptation.}
The executable harness can be used with different backbone models. 
In our main implementation, we adapt a Qwen3.5-27B backbone to better follow tool-using shopping policies. 
The policy is initialized from supervised fine-tuning and further optimized with reinforcement learning over shopping rollouts. 
During training, the model interacts with the same runtime used at inference time, including catalog tools and execution guards. 
Rollouts produce dialogue trajectories, tool traces, and outcome signals, which are scored by task-specific graders. 
Before optimization, we filter groups dominated by infrastructure failures or zero reward variance, and then apply group relative policy optimization. 
This adaptation improves the base policy's ability to follow shopping-specific actions, such as asking useful clarifying questions, retrieving relevant evidence, revising plans after user feedback, and formulating grounded recommendations.

\paragraph{Complementarity between adaptation and executable control.}
Backbone adaptation and executable control address different sources of failure. 
Adaptation improves the model's learned shopping behavior, including action selection and response formulation under evolving user constraints. 
In contrast, the session-control harness imposes explicit structure on the interaction process, preventing the model from ignoring revised constraints, recommending from stale evidence, or making unsupported product claims. 
This complementarity explains why the full \textsc{RealShop\_Agent} outperforms both the prompt-only backbone and the adapted backbone without the complete control loop.

\section{Training Details}
\label{app:training_details}

\subsection{Training Setup}

We adapt the Qwen3.5-27B shopping policy using supervised fine-tuning (SFT) followed by online reinforcement learning.
The SFT stage initializes the model to follow the shopping-agent interaction format, invoke tools correctly, and generate concise grounded responses.
We then optimize the policy with group relative policy optimization (GRPO) using trajectories generated by the current policy in the executable shopping runtime.

During RL training, each prompt is executed through the shopping-agent harness, which exposes the corresponding tools and produces a complete tool-using trajectory.
Task-specific graders assign scalar rewards based on trajectory-level behavior, including recommendation quality, constraint satisfaction, catalog grounding, tool use, and task completion.
The training prompts are disjoint from the held-out \textsc{RealWorldShop} evaluation episodes.

\begin{table}[t]
\centering
\small
\setlength{\tabcolsep}{4pt}
\renewcommand{\arraystretch}{1.08}
\begin{tabular}{l r}
\toprule
\textbf{Item} & \textbf{Value} \\
\midrule
SFT examples & 10K \\
RL prompt seeds & 2K \\
RL training prompts & 1.5K \\
Held-out development prompts & 0.5K \\
GRPO group size & 8 rollouts per prompt \\
Max assistant turns per rollout & 30 \\
\bottomrule
\end{tabular}
\caption{Training scale for backbone adaptation.}
\label{tab:training_scale}
\end{table}

\subsection{Training Tasks and Rewards}

The training mixture contains commerce-related agent tasks covering search and recommendation, routing, service, and order-preview scenarios.
These prompts serve as seeds for online rollout generation rather than supervised input--output pairs.
The training runtime uses catalog and tool interfaces aligned with those used at evaluation time.

Each task-specific grader returns a scalar reward composed of normalized signals.
At a high level, the reward is written as
\[
r =
r_{\mathrm{task}}
+
r_{\mathrm{ground}}
+
r_{\mathrm{constraint}}
+
r_{\mathrm{tool}}
-
p_{\mathrm{invalid}},
\]
where the active terms depend on the task type.
Positive rewards encourage task completion, product relevance, constraint satisfaction, catalog grounding, and correct tool use, while penalties discourage unsupported product claims, invalid tool paths, redundant retrieval, and unsafe transaction behavior.
Rewards are normalized within each prompt group before GRPO advantage computation.

\subsection{GRPO Optimization}

For each prompt, we sample a group of trajectories and compute group-relative advantages as
\[
A_i =
\frac{
r_i-\operatorname{mean}(\{r_j\}_{j=1}^{G})
}{
\operatorname{std}(\{r_j\}_{j=1}^{G})+\epsilon
}.
\]
The policy is optimized with the clipped GRPO objective using the configuration summarized in Table~\ref{tab:training_hyperparameters}.
Rollout groups dominated by infrastructure failures or with zero reward variance are excluded before optimization.

\begin{table}[t]
\centering
\small
\setlength{\tabcolsep}{4pt}
\renewcommand{\arraystretch}{1.10}
\begin{tabular}{@{}p{0.50\linewidth}p{0.34\linewidth}@{}}
\toprule
\textbf{Hyperparameter} & \textbf{Value} \\
\midrule
Learning rate & \texttt{1e-6} \\
Weight decay & \texttt{0.1} \\
GRPO clipping & \texttt{0.2} \\
High clipping threshold & \texttt{0.28} \\
Global batch size & \texttt{64} \\
Samples per prompt & \texttt{8} \\
Rollout temperature & \texttt{1.0} \\
Gradient clipping & \texttt{1.0} \\
KL coefficient & \texttt{0.00} \\
\bottomrule
\end{tabular}
\caption{Main hyperparameters used for GRPO optimization. Additional implementation details follow the SLIME~\cite{slime_github} training framework.}
\label{tab:training_hyperparameters}
\end{table}

\section{Prompt Design}
\label{app:prompt-design}

\begin{figure*}[t]
\centering

\begin{tcolorbox}[
    enhanced,
    colback=black!3,
    colframe=promptblue,
    coltitle=white,
    colbacktitle=promptblue,
    title={Profile Synthesis Prompt},
    fonttitle=\bfseries,
    boxsep=3pt,
    left=4pt,
    right=4pt,
    top=3pt,
    bottom=3pt,
    arc=3pt,
    width=\linewidth
]

\footnotesize
\raggedright
\setlength{\emergencystretch}{2em}

You are generating a structured user profile for a multi-turn e-commerce shopping benchmark.

\smallskip
\textbf{Goal.}
Create a realistic shopping user whose needs can unfold across a full conversation.
The profile should support evaluation of clarification, state tracking, constraint updating,
multi-intent coordination, bundle/split-order planning, and grounded recommendation.

\smallskip
\textbf{Input fields.}
You will receive:
\begin{itemize}[
    leftmargin=*,
    itemsep=0.3pt,
    topsep=0.7pt,
    parsep=0pt,
    partopsep=0pt
]
    \item product category anchors and source category paths;

    \item a shopping scenario describing the user's real-world context;

    \item a session plan specifying whether the intent is explicit or ambiguous,
    whether the session contains single or multiple subgoals, and when hidden constraints,
    preference updates, rejections, comparisons, confirmations, or intent shifts may appear;

    \item a scenario archetype, such as
    \texttt{combo\_from\_scratch},
    \texttt{combo\_core\_upgrade},
    \texttt{replacement\_repair},
    \texttt{scenario\_bundle},
    \texttt{consumable\_restock},
    \texttt{vague\_exploration},
    \texttt{mixed\_errands}, or
    \texttt{urgent\_timing};

    \item constraint modifiers, such as
    \texttt{compatibility\_sensitive},
    \texttt{budget\_tradeoff},
    \texttt{high\_risk\_decision},
    \texttt{logistics\_time\_sensitive}, or
    \texttt{bundle\_split\_decision};

    \item optional catalog evidence, including product titles, brands, categories,
    prices, attributes, and fulfillment tags.
\end{itemize}

\smallskip
\textbf{Generation requirements.}
Generate a profile that satisfies the following requirements:
\begin{itemize}[
    leftmargin=*,
    itemsep=0.3pt,
    topsep=0.7pt,
    parsep=0pt,
    partopsep=0pt
]
    \item Separate long-term user preferences from current session-specific needs.

    \item Specify whether the session is single-intent or multi-intent consistently
    with the provided session plan.

    \item If the session is multi-intent, list subgoals, priority order, must-finish items,
    deferrable items, and bundle/split-order preference consistently with the provided
    session plan.

    \item Distinguish hard constraints, soft preferences, and negotiable constraints.

    \item Include compatibility, size, delivery, budget, safety, or fulfillment constraints
    when they are implied by the scenario.

    \item Include an interaction style with coping style, interaction mode, verbosity,
    politeness, skepticism, decisiveness, patience, and comparison habit.

    \item Generate the latent intent, constraints, interaction style, disclosure plan,
    and multi-intent structure consistently with the provided session plan.

    \item Do not modify, rewrite, or contradict the provided session plan.

    \item Include an evaluation focus describing what the shopping agent must do correctly
    to succeed.
\end{itemize}

\smallskip
\textbf{Ambiguous-intent profiles.}
If the profile type is \texttt{ambiguous\_intent}, do not expose the full target item
in the opening utterance. Instead:
\begin{itemize}[
    leftmargin=*,
    itemsep=0.3pt,
    topsep=0.7pt,
    parsep=0pt,
    partopsep=0pt
]
    \item write an opening utterance that reveals only a scenario, problem, recipient,
    or desired outcome;

    \item store the true goal in \texttt{latent\_intent};

    \item store hidden constraints in
    \texttt{latent\_intent.must\_have\_constraints} and
    \texttt{latent\_intent.nice\_to\_have\_constraints};

    \item construct a \texttt{disclosure\_plan} consistently with the provided session plan,
    specifying how the corresponding hidden intent, constraints, preference updates,
    rejections, or intent shifts are revealed over turns.
\end{itemize}

\smallskip
\textbf{Catalog-grounding rules.}
If catalog evidence is provided:
\begin{itemize}[
    leftmargin=*,
    itemsep=0.3pt,
    topsep=0.7pt,
    parsep=0pt,
    partopsep=0pt
]
    \item concrete brands, attributes, categories, fulfillment tags, and product-specific
    constraints should be grounded in the evidence;

    \item unsupported details must not be written as hard constraints;

    \item if a detail is plausible but not supported by evidence, write it as a soft
    or negotiable preference;

    \item avoid inventing platform services, delivery guarantees, certifications,
    or exact product attributes that are not present in the evidence.
\end{itemize}

\smallskip
\textbf{Output format.}
Return valid JSON with the following fields:
\begin{itemize}[
    leftmargin=*,
    itemsep=0.3pt,
    topsep=0.7pt,
    parsep=0pt,
    partopsep=0pt
]
    \item \texttt{profile\_id},
    \texttt{profile\_type},
    \texttt{persona\_seed};

    \item \texttt{purchase\_intent\_scope},
    \texttt{purchase\_intent\_subgoals},
    \texttt{multi\_intent\_plan};

    \item \texttt{shopping\_scenario},
    \texttt{synthesis\_anchor};

    \item \texttt{basic\_information},
    \texttt{long\_term\_preference},
    \texttt{historical\_behavior};

    \item \texttt{current\_intent},
    \texttt{constraints},
    \texttt{interaction\_style};

    \item \texttt{preference\_priority},
    \texttt{risk\_preference},
    \texttt{urgency\_level};

    \item \texttt{initial\_utterance},
    \texttt{initial\_utterance\_style},
    \texttt{latent\_intent},
    \texttt{disclosure\_plan},
    \texttt{evaluation\_focus};

    \item \texttt{category\_specific\_extension}.
\end{itemize}

\end{tcolorbox}

\vspace{-0.5em}

\caption{
Profile synthesis prompt used to construct structured user profiles for
\textsc{RealWorldShop}. }
\label{fig:profile_synthesis_prompt}

\end{figure*}

\clearpage

\begin{figure*}[t]
\centering

\begin{tcolorbox}[
    enhanced,
    colback=black!3,
    colframe=promptblue,
    coltitle=white,
    colbacktitle=promptblue,
    title={User Simulator Prompt},
    fonttitle=\bfseries,
    boxsep=3pt,
    left=4pt,
    right=4pt,
    top=3pt,
    bottom=3pt,
    arc=3pt,
    width=\linewidth
]

\footnotesize
\raggedright
\setlength{\emergencystretch}{2em}

You are playing the user in a controlled multi-turn e-commerce shopping conversation.

\smallskip
\textbf{Goal.}
Generate the next user utterance according to the structured user profile, hidden state,
dialogue history, and target user action.
The utterance should be natural, but it must remain faithful to the profile and the assigned action.

\smallskip
\textbf{Input fields.}
You will receive:
\begin{itemize}[
    leftmargin=*,
    itemsep=0.3pt,
    topsep=0.7pt,
    parsep=0pt,
    partopsep=0pt
]
    \item the visible and hidden user profile;

    \item the user's long-term preferences and current session intent;

    \item hard, soft, and negotiable constraints;

    \item latent intent and disclosure plan for ambiguous profiles;

    \item dialogue history;

    \item the target user action for this turn;

    \item interaction style, including patience, skepticism, verbosity, decisiveness,
    politeness, and comparison habit.
\end{itemize}

\smallskip
\textbf{Action control.}
Follow the target action exactly:
\begin{itemize}[
    leftmargin=*,
    itemsep=0.3pt,
    topsep=0.7pt,
    parsep=0pt,
    partopsep=0pt
]
    \item \textsc{Provide}: answer the assistant's clarification question or provide a missing slot.

    \item \textsc{Refine}: make an existing requirement more specific without changing it.

    \item \textsc{Update}: change a previously stated constraint, budget, preference,
    priority, active subgoal, or shopping intent.

    \item \textsc{Override}: let the current session need override a long-term preference.

    \item \textsc{Reject}: reject a recommendation and give a profile-consistent reason.

    \item \textsc{Compare}: ask the assistant to compare candidates or trade-offs.

    \item \textsc{Confirm}: accept a recommendation or purchase plan if it satisfies
    the active constraints.

    \item \textsc{End}: terminate the session without convergence because the user is
    impatient, unconvinced, or no longer wishes to continue.
\end{itemize}

\smallskip
\textbf{Hidden-state rules.}
Do not reveal the full hidden profile at once.
\begin{itemize}[
    leftmargin=*,
    itemsep=0.3pt,
    topsep=0.7pt,
    parsep=0pt,
    partopsep=0pt
]
    \item Reveal hidden constraints only when the assistant asks a relevant question,
    when the target action requires it, or when a recommendation violates them.

    \item For ambiguous-intent sessions, begin from the opening style and avoid naming
    all target products immediately.

    \item If the assistant asks a good clarification question, reveal the corresponding
    slot from the disclosure plan.

    \item If the assistant recommends an item that violates a hard constraint, reject it
    and restate the violated constraint.

    \item If the assistant forgets a previously revealed constraint, respond according
    to the user's patience and skepticism.
\end{itemize}

\smallskip
\textbf{Style rules.}
Realize the utterance using the user's interaction style:
\begin{itemize}[
    leftmargin=*,
    itemsep=0.3pt,
    topsep=0.7pt,
    parsep=0pt,
    partopsep=0pt
]
    \item low-patience users should be concise and may end early after repeated failures;

    \item high-skepticism users should ask for evidence, compatibility, price,
    or fulfillment details;

    \item efficiency-first users should prefer direct answers and dislike redundant questions;

    \item collaborative users may provide extra context and tolerate clarification;

    \item comparison-oriented users should ask for trade-offs among candidates.
\end{itemize}

\smallskip
\textbf{Output constraints.}
Return only one user utterance.
Do not include analysis, hidden fields, action labels, JSON, or explanations.

\end{tcolorbox}

\vspace{-0.5em}

\caption{
User simulator prompt used to generate controlled multi-turn user behavior in
\textsc{RealWorldShop}. 
}
\label{fig:user_simulator_prompt}

\end{figure*}

\clearpage

\begin{figure*}[t]
\centering

\begin{tcolorbox}[
    enhanced,
    colback=black!3,
    colframe=promptblue,
    coltitle=white,
    colbacktitle=promptblue,
    title={Shopping-Agent Prompt},
    fonttitle=\bfseries,
    boxsep=3pt,
    left=4pt,
    right=4pt,
    top=3pt,
    bottom=3pt,
    arc=3pt,
    width=\linewidth
]

\footnotesize
\raggedright
\setlength{\emergencystretch}{2em}

You are a conversational shopping assistant for a real-world e-commerce platform.

\smallskip
\textbf{Goal.}
Help the user reach a grounded and actionable purchase decision over a multi-turn conversation.
You should clarify hidden needs, track evolving constraints, retrieve product evidence,
compare candidates, and recommend only when the recommendation is supported by current
user state and catalog evidence.

\smallskip
\textbf{Visible information.}
You can observe:
\begin{itemize}[
    leftmargin=*,
    itemsep=0.3pt,
    topsep=0.7pt,
    parsep=0pt,
    partopsep=0pt
]
    \item the dialogue history;

    \item the visible user prior, if available;

    \item catalog search and product tools;

    \item previous tool results and candidate products.
\end{itemize}

You cannot observe the user's hidden constraints, latent intent, disclosure plan,
or evaluation labels.

\smallskip
\textbf{Session-state requirements.}
Maintain an internal session state containing:
\begin{itemize}[
    leftmargin=*,
    itemsep=0.3pt,
    topsep=0.7pt,
    parsep=0pt,
    partopsep=0pt
]
    \item active shopping intent and active subgoals;

    \item hard constraints, soft preferences, and negotiable constraints;

    \item rejected products and rejection reasons;

    \item confirmed requirements;

    \item unresolved uncertainties;

    \item stale candidates whose assumptions no longer match the current state.
\end{itemize}

\smallskip
\textbf{Decision policy.}
At each turn, choose the next behavior according to the current state:
\begin{itemize}[
    leftmargin=*,
    itemsep=0.3pt,
    topsep=0.7pt,
    parsep=0pt,
    partopsep=0pt
]
    \item \textsc{Clarify} if a blocking field is missing, such as budget, size,
    compatibility, recipient, urgency, or usage scenario.

    \item \textsc{Retrieve} if the user's intent is clear but no catalog evidence
    has been collected.

    \item \textsc{Verify} if candidates exist but have not been checked against
    hard constraints.

    \item \textsc{Compare} if the user asks about trade-offs or if multiple viable
    candidates remain.

    \item \textsc{Recommend} only when candidates are grounded, compatible with
    hard constraints, and aligned with the active subgoal.

    \item \textsc{Confirm} when the user is ready to commit and the recommendation
    is actionable.
\end{itemize}

\smallskip
\textbf{Constraint-handling rules.}
\begin{itemize}[
    leftmargin=*,
    itemsep=0.3pt,
    topsep=0.7pt,
    parsep=0pt,
    partopsep=0pt
]
    \item Hard constraints must not be violated.

    \item When the user updates a hard constraint, invalidate candidates that
    no longer satisfy it.

    \item Session-specific constraints override long-term preferences.

    \item In multi-intent sessions, keep must-finish and deferrable subgoals separate.

    \item In bundle or split-order sessions, explain which items should be bought
    together and which should be ordered separately.

    \item If the user rejects an item, do not recommend the same item again unless
    the user explicitly reconsiders it.
\end{itemize}

\smallskip
\textbf{Grounding rules.}
\begin{itemize}[
    leftmargin=*,
    itemsep=0.3pt,
    topsep=0.7pt,
    parsep=0pt,
    partopsep=0pt
]
    \item Use catalog tools before making SKU-level claims.

    \item Do not invent product attributes, compatibility, price, inventory,
    delivery time, certification, or fulfillment services.

    \item If catalog evidence is incomplete, state uncertainty and ask for
    missing information.

    \item Product explanations should cite concrete evidence from retrieved records,
    such as category, attributes, price, fulfillment tag, or compatibility field.
\end{itemize}

\smallskip
\textbf{Response style.}
Be concise, helpful, and action-oriented.
Avoid over-questioning when enough information is available.
Avoid premature recommendation when blocking constraints are missing.
For low-patience users, ask at most one targeted clarification question before moving forward.

\end{tcolorbox}

\vspace{-0.5em}

\caption{
Shopping-agent prompt used to control session-level decision making in
\textsc{RealWorldShop}.
}
\label{fig:shopping_agent_prompt}

\end{figure*}

\clearpage

\begin{figure*}[t]
\centering

\begin{tcolorbox}[
    enhanced,
    colback=black!3,
    colframe=promptblue,
    coltitle=white,
    colbacktitle=promptblue,
    title={Evaluation Judge Prompt},
    fonttitle=\bfseries,
    boxsep=3pt,
    left=4pt,
    right=4pt,
    top=3pt,
    bottom=3pt,
    arc=3pt,
    width=\linewidth
]

\footnotesize
\raggedright
\setlength{\emergencystretch}{2em}

You are evaluating a completed multi-turn shopping-agent dialogue.

\smallskip
\textbf{Goal.}
Score whether the assistant handled the shopping session correctly.
Do not reward fluent but unsupported responses.
Focus on whether the assistant understood the user, tracked state, updated constraints,
used product evidence, and guided the session toward an actionable purchase decision.

\smallskip
\textbf{Input fields.}
You will receive:
\begin{itemize}[
    leftmargin=*,
    itemsep=0.3pt,
    topsep=0.7pt,
    parsep=0pt,
    partopsep=0pt
]
    \item structured episode profile;

    \item visible and hidden user constraints;

    \item dialogue trajectory;

    \item tool trace;

    \item recommended products or final purchase plan;

    \item termination reason.
\end{itemize}

\smallskip
\textbf{Turn-level scoring.}
For each assistant turn, score the following dimensions on a $\{0,1,2\}$ scale:
\begin{itemize}[
    leftmargin=*,
    itemsep=0.3pt,
    topsep=0.7pt,
    parsep=0pt,
    partopsep=0pt
]
    \item \textbf{Need Understanding}: whether the assistant correctly understands
    the current user request, newly revealed constraints, and changes from previous turns.

    \item \textbf{Recommendation Accuracy}: whether recommended products are appropriate,
    grounded, and consistent with known constraints.

    \item \textbf{Rationale Quality}: whether recommendation reasons are specific,
    evidence-based, and useful for the user's decision.
\end{itemize}

Use N/A for recommendation-related dimensions when no recommendation is made.

\smallskip
\textbf{Session-level scoring.}
Score the full trajectory on a $\{0,1,2\}$ scale:
\begin{itemize}[
    leftmargin=*,
    itemsep=0.3pt,
    topsep=0.7pt,
    parsep=0pt,
    partopsep=0pt
]
    \item \textbf{Clarification}: asks about key missing fields before reliable recommendation.

    \item \textbf{State Tracking}: remembers revealed needs, constraints, priorities,
    and rejected items.

    \item \textbf{Constraint Updating}: revises behavior after the user changes
    or overrides a requirement.

    \item \textbf{Decision Pacing}: chooses when to clarify, retrieve, verify, compare,
    recommend, or confirm without premature commitment or excessive delay.

    \item \textbf{Personalized Recommendation}: aligns final or near-final recommendations
    with the user's profile and session-specific constraints.
\end{itemize}

\smallskip
\textbf{Holistic outcome.}
Assign:
\begin{itemize}[
    leftmargin=*,
    itemsep=0.3pt,
    topsep=0.7pt,
    parsep=0pt,
    partopsep=0pt
]
    \item \texttt{overall\_success}:
    \texttt{success}, \texttt{partial}, or \texttt{failed};

    \item \texttt{convergence\_quality}:
    \texttt{strong}, \texttt{moderate}, \texttt{weak}, or \texttt{none}.
\end{itemize}

\smallskip
\textbf{Failure criteria.}
Mark the session down if the assistant:
\begin{itemize}[
    leftmargin=*,
    itemsep=0.3pt,
    topsep=0.7pt,
    parsep=0pt,
    partopsep=0pt
]
    \item recommends before clarifying blocking constraints;

    \item forgets or contradicts previously revealed information;

    \item continues using stale candidates after a constraint update;

    \item violates hard constraints;

    \item makes unsupported SKU-level, compatibility, price, or delivery claims;

    \item fails to separate urgent and deferrable subgoals;

    \item over-questions a low-patience user or fails to converge when sufficient
    information is available.
\end{itemize}

\smallskip
\textbf{Output format.}
Return structured JSON with turn-level scores, session-level scores, holistic outcome,
convergence quality, evidence turns, and concise justifications.

\end{tcolorbox}

\vspace{-0.5em}

\caption{
Evaluation judge prompt used to assess completed shopping-agent trajectories in
\textsc{RealWorldShop}. 
}
\label{fig:evaluation_judge_prompt}

\end{figure*}

\clearpage

\section{Case Study}
\label{app:case-study}

We provide representative case studies to illustrate how \textsc{RealWorldShop} stresses session-level decision control. 
The selected profiles cover compatibility-sensitive shopping, ambiguous intent, multi-intent coordination, low-patience users, and split-order decisions. 
The failure cases highlight common weaknesses observed in baseline agents, including missing clarification, stale-state recommendations, priority loss, and unsupported compatibility claims.

\begin{table*}[t]
\centering
\small
\setlength{\tabcolsep}{4pt}
\renewcommand{\arraystretch}{1.12}
\begin{tabularx}{\textwidth}{@{}p{0.18\textwidth}p{0.24\textwidth}p{0.26\textwidth}X@{}}
\toprule
\textbf{Case} & \textbf{Scenario} & \textbf{Stress Factor} & \textbf{Diagnostic Target} \\
\midrule
Profile A & Compatibility-sensitive multi-intent shopping & Old-wall-box fit, narrow balcony space, urgent vs. deferrable items & State tracking, compatibility verification, split-order planning \\
Profile B & Ambiguous core-item lighting upgrade & Vague study-room lighting need; 75\,mm opening and budget priority revealed later & Clarification, hidden-constraint elicitation, hard-constraint propagation, candidate verification \\
Profile C & Low-patience family purchase & Speaker compatibility, fast convergence, later add-on item & Verification before recommendation, low-patience decision pacing \\
\midrule
Failure 1 & Missing clarification & Underspecified small-kitchen request & Avoid premature recommendation under hidden constraints \\
Failure 2 & Priority loss in multi-intent shopping & Urgent setup items vs. optional decorative items & Preserve subgoal priority and avoid stale planning \\
Failure 3 & Constraint update not propagated & 85\,mm recommendation after user corrects to 75\,mm & Invalidate stale candidates after hard-constraint update \\
Failure 4 & Unsupported compatibility claim & 6.35\,mm speaker connection for a digital piano & Avoid ungrounded product claims and verify catalog evidence \\
\bottomrule
\end{tabularx}
\caption{Summary of representative case studies and diagnostic targets.}
\label{tab:case_study_summary}
\end{table*}

\subsection{Representative Profiles}
\label{sec:representative_profiles}

To illustrate the diversity and controllability of \textsc{RealWorldShop} profiles, we present three representative cases. 
Our benchmark contains both \textit{explicit-intent} profiles, where the initial request already exposes directly searchable goals, and \textit{ambiguous-intent} profiles, where key goals or constraints are revealed only through interaction. 
Each profile follows a normalized schema that separates \textit{hard}, \textit{soft}, and \textit{negotiable} requirements, covering budget, compatibility, risk tolerance, logistics, brand preference, and interaction style. 
To reduce unverifiable assumptions, profile synthesis can be conditioned on local catalog evidence; constraints that cannot be grounded in catalog attributes are downgraded to soft or negotiable preferences. 
All profiles are programmatically normalized and validated for schema consistency, multi-intent structure, ambiguous-intent fields, category anchoring, and catalog grounding. 
Further details are provided in Appendix~\ref{app:benchmark-profile}.

\begin{figure*}[t]
\begin{tcolorbox}[
  enhanced, breakable,
  colback=black!3, colframe=promptblue,
  coltitle=white, colbacktitle=promptblue,
  title={Profile A: Compatibility-Sensitive Multi-Intent Shopping},
  fonttitle=\bfseries, fontupper=\footnotesize,
  boxsep=4pt, left=5pt,right=5pt,top=4pt,bottom=4pt, arc=3pt
]
\begin{tabularx}{\textwidth}{@{}p{0.17\textwidth}X@{}}
\textbf{Profile type} & Partially explicit multi-intent profile with latent compatibility and logistics constraints. \\
\textbf{Persona} & A specification-conscious, price-sensitive user buying home-renovation items for elderly parents. \\
\textbf{Scenario} & The user wants to build a balcony wellness corner involving a leakage-protection switch, narrow planters, mild tea, and a compact bathtub. \\
\textbf{Category anchors} & Electrical safety device, balcony planter, mild tea product, and compact bathtub. \\
\textbf{Session plan} & The user first asks for the switch and planters, then reveals old-wall-box and narrow-door constraints. After seeing candidate products, the user asks whether urgent and deferrable items should be split into separate orders. \\
\textbf{Requirement schema} & 
\textit{Hard}: the switch must fit an old wall box; the bathtub must pass through a narrow bathroom door; urgent electrical safety items should be ordered first. 
\textit{Soft}: price sensitivity, concise comparison, and mild tea preference. 
\textit{Negotiable}: planters, tea, and bathtub can be postponed or ordered separately. \\
\textbf{Interaction style} & Efficiency-first, medium skepticism, concise comparison expected. \\
\textbf{Grounding and validation} & Compatibility and logistics constraints should be verified against product specifications or catalog attributes when available; unsupported compatibility assumptions are treated as soft preferences rather than hard constraints. \\
\textbf{Evaluation focus} & Preserve compatibility constraints across turns, prioritize the urgent safety item, and produce a reasonable split-order strategy. \\
\end{tabularx}
\end{tcolorbox}
\caption{Representative profile for compatibility-sensitive multi-intent shopping.}
\label{fig:profile_case_a}
\end{figure*}

\begin{figure*}[t]
\centering
\begin{tcolorbox}[
  enhanced,
  colback=black!3, colframe=promptblue,
  coltitle=white, colbacktitle=promptblue,
  title={Profile B: Ambiguous Core-Item Upgrade},
  fonttitle=\bfseries,
  fontupper=\scriptsize,
  boxsep=3pt,
  left=4pt,right=4pt,top=3pt,bottom=3pt,
  arc=3pt
]
\renewcommand{\arraystretch}{1.02}
\begin{tabularx}{\textwidth}{@{}>{\bfseries\raggedright\arraybackslash}p{0.145\textwidth}@{\hspace{0.8em}}>{\raggedright\arraybackslash}X@{}}

Profile type &
Ambiguous-intent profile centered on a high-stakes home-improvement need with gradually revealed installation and budget constraints. \\

Persona &
A budget-conscious father improving his child's study-room environment. \\

Scenario &
The user wants more comfortable study-room lighting for reading and drawing, while also considering low-priority drawing tools and household supplies. \\

Initial utterance &
``I want to improve the lighting in my child's study room. It should be more comfortable for reading and drawing. What would you recommend?'' \\

Category anchors &
Study-room lighting, drawing tools, and household supplies. \\

Session plan &
The user starts with a vague lighting need. If the agent probes the setup, he reveals a preference for a no-main-light design. If asked about installation, he reveals that the existing ceiling opening is 75\,mm and cannot be re-cut. If asked about quality or budget, he prioritizes glare control, high color rendering, and allocating most of the budget to the core lighting item. \\

Requirement schema &
\textit{Hard}: the final lighting solution must fit the revealed 75\,mm opening and support comfortable study use.
\textit{Soft}: anti-glare design, high color rendering, and cost-effectiveness.
\textit{Negotiable}: drawing tools and household supplies are peripheral and lower priority. \\

Interaction style &
Directive, high skepticism, low patience, key-option comparison preferred. \\

Grounding &
Installation-sensitive claims must be grounded in catalog specifications such as cutout size, glare-control design, beam angle, and color-rendering index. Products lacking these attributes should not be treated as verified matches. \\

Evaluation focus &
Elicit the true product target, clarify installation compatibility, propagate newly revealed hard constraints, and recover after incompatible recommendations. \\

\end{tabularx}
\end{tcolorbox}
\vspace{-0.6em}
\caption{Representative profile for an ambiguous core-item upgrade scenario.}
\label{fig:profile_case_b}
\end{figure*}

\begin{figure*}[t]
\begin{tcolorbox}[
  enhanced, breakable,
  colback=black!3, colframe=promptblue,
  coltitle=white, colbacktitle=promptblue,
  title={Profile C: Low-Patience Compatibility-Sensitive Family Purchase},
  fonttitle=\bfseries, fontupper=\footnotesize,
  boxsep=4pt, left=5pt,right=5pt,top=4pt,bottom=4pt, arc=3pt
]
\begin{tabularx}{\textwidth}{@{}p{0.17\textwidth}X@{}}
\textbf{Profile type} & Compatibility-sensitive family purchase with low patience and a direct-answer preference. \\
\textbf{Persona} & A pragmatic parent buying a better digital piano for a child after an exam milestone. \\
\textbf{Scenario} & The user wants a professional digital piano, needs to verify compatibility with an old external speaker, and may add an adapter cable and small learning accessories. \\
\textbf{Initial utterance} & ``I want to replace my child's entry-level keyboard with a professional digital piano. Can it connect to our old speaker with a normal 6.35mm audio cable?'' \\
\textbf{Category anchors} & Professional digital piano, external speaker connection, audio cable or adapter, and learning accessories. \\
\textbf{Session plan} & The user first asks about compatibility. If the agent clarifies the speaker interface, the user confirms the 6.35mm requirement. If the agent recommends products before checking compatibility, the user becomes skeptical and asks for a direct answer. \\
\textbf{Requirement schema} & 
\textit{Hard}: the piano must support the required audio-output pathway or a verified compatible adapter; unsupported compatibility claims are unacceptable. 
\textit{Soft}: professional quality, child-learning suitability, and reasonable trade-offs between function and price. 
\textit{Negotiable}: adapter cables and small learning accessories can be postponed if the core piano choice is not yet confirmed. \\
\textbf{Interaction style} & Low patience, medium skepticism, trade-off seeking, concise answers preferred. \\
\textbf{Grounding and validation} & Audio-interface compatibility should be checked against product specifications, port descriptions, or verified adapter requirements; if catalog evidence is missing, the agent should explicitly ask for the speaker model or avoid making a definitive claim. \\
\textbf{Evaluation focus} & Verify compatibility before recommendation, avoid unsupported claims, and converge quickly under low-patience pressure. \\
\end{tabularx}
\end{tcolorbox}
\caption{Representative profile for low-patience compatibility-sensitive purchase.}
\label{fig:profile_case_c}
\end{figure*}

\subsection{Representative Failure Modes}

\begin{figure*}[t]
\begin{tcolorbox}[
  enhanced, breakable,
  colback=red!2, colframe=red!55!black,
  coltitle=white, colbacktitle=red!65!black,
  title={Failure Case 1: Missing Clarification},
  fonttitle=\bfseries, fontupper=\footnotesize,
  boxsep=4pt, left=5pt,right=5pt,top=4pt,bottom=4pt, arc=3pt
]
\begin{tabularx}{\textwidth}{@{}p{0.16\textwidth}X@{}}
\textbf{Scenario} & The user gives an underspecified small-kitchen request. \\
\textbf{Dialogue} &
\textbf{User:} I want something useful for my small kitchen. It should make cooking easier. \newline
\textbf{Assistant:} I recommend a large multifunctional air fryer with high capacity and many cooking modes. \newline
\textbf{User:} My counter is tiny, I live alone, and I care about noise. I did not ask for a large appliance. \\
\textbf{Failure} & The assistant recommends before eliciting hidden constraints such as space, household size, noise tolerance, and preferred appliance type. \\
\textbf{Expected} & Ask targeted clarification questions before retrieving or recommending products. \\
\end{tabularx}
\end{tcolorbox}
\caption{Representative failure caused by missing clarification under ambiguous intent.}
\label{fig:failure_missing_clarification}
\end{figure*}

\begin{figure*}[t]
\centering
\begin{tcolorbox}[
  enhanced,
  colback=red!2,
  colframe=red!55!black,
  coltitle=white,
  colbacktitle=red!65!black,
  title={Failure Case 2: Priority Loss in Multi-Intent Shopping},
  fonttitle=\bfseries,
  fontupper=\footnotesize,
  boxsep=4pt,
  left=5pt,
  right=5pt,
  top=4pt,
  bottom=4pt,
  arc=3pt,
  width=\textwidth
]
\begin{tabularx}{\linewidth}{@{}p{0.16\linewidth}X@{}}
\textbf{Scenario} 
& The user has urgent opening-day items and deferrable decorative items. \\

\textbf{Dialogue} 
& \textbf{User:} I need a commercial display fridge, bottled drinks for opening day, a flatbed trolley, and some decorative items. \par
  \textbf{Assistant:} I suggest buying all of them together to maximize discounts. \par
  \textbf{User:} The fridge, drinks, and trolley are opening-day essentials. Decorations can wait. \par
  \textbf{Assistant:} Then I recommend starting with decorative shelves and matching wall art to improve the store atmosphere. \\

\textbf{Failure} 
& The assistant loses the revealed priority structure and shifts from urgent operational subgoals to optional decorative items. \\

\textbf{Expected} 
& Track urgent and deferrable subgoals separately, prioritize the fridge, drinks, and trolley, and recommend a split-order plan when optional items can be postponed. \\
\end{tabularx}
\end{tcolorbox}
\caption{Representative failure caused by priority loss in a multi-intent shopping session.}
\label{fig:failure_priority_loss}
\end{figure*}

\begin{figure*}[t]
\centering
\begin{tcolorbox}[
  enhanced,
  colback=red!2,
  colframe=red!55!black,
  coltitle=white,
  colbacktitle=red!65!black,
  title={Failure Case 3: Constraint Update Not Propagated},
  fonttitle=\bfseries,
  fontupper=\footnotesize,
  boxsep=4pt,
  left=5pt,
  right=5pt,
  top=4pt,
  bottom=4pt,
  arc=3pt,
  width=\textwidth
]
\begin{tabularx}{\linewidth}{@{}p{0.16\linewidth}X@{}}
\textbf{Scenario} 
& The user reveals a hard installation constraint after an unsuitable recommendation. \\

\textbf{Dialogue} 
& \textbf{User:} I want to improve the lighting in my child's study room. It should be comfortable for reading and drawing. \par
  \textbf{Assistant:} I found a recessed spotlight with an 85\,mm cutout. It has anti-glare design and high color rendering. \par
  \textbf{User:} 85\,mm will not work. The existing ceiling opening is 75\,mm, and I cannot re-cut the ceiling. \par
  \textbf{Assistant:} Understood. This 85\,mm spotlight is still a good option because it has better brightness. \\

\textbf{Failure} 
& The assistant acknowledges the newly revealed hard constraint but continues recommending from a stale candidate set. \\

\textbf{Expected} 
& Invalidate the stale candidate and retrieve or verify products compatible with the revealed 75\,mm opening. \\
\end{tabularx}
\end{tcolorbox}
\caption{Representative failure caused by stale candidate reuse after a newly revealed hard constraint.}
\label{fig:failure_constraint_update}
\end{figure*}

\begin{figure*}[t]
\centering
\begin{tcolorbox}[
  enhanced,
  colback=red!2,
  colframe=red!55!black,
  coltitle=white,
  colbacktitle=red!65!black,
  title={Failure Case 4: Unsupported Compatibility Claim},
  fonttitle=\bfseries,
  fontupper=\footnotesize,
  boxsep=4pt,
  left=5pt,
  right=5pt,
  top=4pt,
  bottom=4pt,
  arc=3pt,
  width=\textwidth
]
\begin{tabularx}{\linewidth}{@{}p{0.16\linewidth}X@{}}
\textbf{Scenario} 
& The user asks whether a candidate product is compatible with existing equipment. \\

\textbf{Dialogue} 
& \textbf{User:} I want to buy a professional digital piano. Can it connect to our old speaker with a normal 6.35mm audio cable? \par
  \textbf{Assistant:} This piano has weighted keys and realistic sound, so I recommend it. \par
  \textbf{User:} But can this model connect with a 6.35mm cable? \par
  \textbf{Assistant:} Yes, this model should work with a 6.35mm cable. Most digital pianos support external speakers. \\

\textbf{Failure} 
& The assistant asserts SKU-level compatibility without verifying the product interface or asking for the speaker model. \\

\textbf{Expected} 
& Check catalog evidence for the piano's output ports, ask for the speaker's input interface when needed, and state uncertainty if compatibility cannot be verified. \\
\end{tabularx}
\end{tcolorbox}
\caption{Representative failure caused by an unsupported product-compatibility claim.}
\label{fig:failure_compatibility_claim}
\end{figure*}

\end{document}